\documentclass[]{InspireOmni_template}

\usepackage[toc,page,header]{appendix}

\usepackage{hyperref}
\usepackage{url}

\usepackage{caption}

\usepackage{algorithm}
\usepackage{algpseudocode}
\usepackage{multirow}
\usepackage{enumitem}
\usepackage{booktabs}
\usepackage{xcolor}
\usepackage{graphicx}
\usepackage{wrapfig}
\usepackage{pdflscape}
\usepackage{subcaption}
\usepackage[table]{xcolor}
\usepackage{etoc}
\definecolor{matchagreen}{HTML}{DDE9D8}
\usepackage{booktabs}
\usepackage{array}
\usepackage{graphicx}
\usepackage[table]{xcolor}
\usepackage{amssymb}

\definecolor{AvgBg}{HTML}{F0F3F7}
\definecolor{OursBg}{HTML}{E7EFE3}
\definecolor{GroupText}{HTML}{596573}

\newcolumntype{L}[1]{>{\raggedright\let\newline\\\arraybackslash\hspace{0pt}}m{#1}}
\newcolumntype{R}[1]{>{\raggedleft\let\newline\\\arraybackslash\hspace{0pt}}m{#1}}

\newcommand{\ignore}[1]{}

\makeatletter
\DeclareRobustCommand\onedot{\futurelet\@let@token\@onedot}
\def\@onedot{\ifx\@let@token.\else.\null\fi\xspace}

\makeatother

\definecolor{MyBlue}{rgb}{0.46, 0.50, 0.61}
\definecolor{MyDarkBlue}{rgb}{0,0.08,0.8}
\definecolor{MyDarkGreen}{RGB}{45,155,45}
\definecolor{MyDarkRed}{rgb}{0.8,0.02,0.02}
\definecolor{MyOrange}{rgb}{1.0, 0.4, 0.2}
\definecolor{MyPurple}{RGB}{111,0,255}
\definecolor{MyRed}{rgb}{0.8,0.0,0.0}
\definecolor{MyGold}{rgb}{0.75,0.6,0.12}
\definecolor{MyDarkgray}{rgb}{0.66, 0.66, 0.66}
\definecolor{MyBrown}{rgb}{0.65, 0.16, 0.16}
\definecolor{MyMutedRose}{rgb}{0.58, 0.29, 0.35}
\definecolor{JiayuanColor}{rgb}{0.60,0.43,0.48}
\definecolor{erranColor}{rgb}{24, 40, 113}

\definecolor{citecolor}{HTML}{696FAD}

\definecolor{bggray}{HTML}{F5F5F5}
\definecolor{pvdblue}{HTML}{DAE8FC}
\definecolor{RoseQuartzBg}{HTML}{F7CAC9}
\definecolor{RoseQuartz}{HTML}{F5A798}
\definecolor{Serenity}{HTML}{92A8D1}
\definecolor{OrangeRed}{rgb}{1.0, 0.27, 0.0}
\definecolor{RoyalBlue}{cmyk}{1, 0.50, 0, 0}
\definecolor{Turquoise}{HTML}{0F4C81}
\definecolor{mint}{rgb}{0.24, 0.71, 0.54}
\definecolor{green}{rgb}{0.0, 0.120, 0.0}

\newdimen\abovecrulesep
\newdimen\belowcrulesep
\makeatletter
\patchcmd{\@@@cmidrule}{\aboverulesep}{\abovecrulesep}{}{}
\patchcmd{\@xcmidrule}{\belowrulesep}{\belowcrulesep}{}{}
\makeatother

\definecolor{mybluetitle}{HTML}{4B527E} %

\definecolor{codegreen}{HTML}{478058}%
\definecolor{codegray}{rgb}{0.5,0.5,0.5}
\definecolor{codepurple}{HTML}{4F5E80} %
\definecolor{backcolour}{rgb}{0.95,0.95,0.92}
\lstdefinestyle{mystyle}{
    backgroundcolor=\color{backcolour},
    commentstyle=\color{codegreen},
    keywordstyle=\color{magenta},
    numberstyle=\tiny\color{codegray},
    stringstyle=\color{codepurple},
    basicstyle=\ttfamily\scriptsize,
    breakatwhitespace=false,
    breaklines=true,
    captionpos=b,
    keepspaces=true,
    frame=none,
    numbersep=5pt,
    showspaces=false,
    showstringspaces=false,
    showtabs=false,
    tabsize=2
}

\newtcolorbox{promptbox}[2][]{
    enhanced, 
    breakable,
    center title,
    left*=0pt, right*=0pt,
    boxsep=2pt, left=5pt, right=5pt,
    skin first=enhanced,
    skin middle=enhanced,
    skin last=enhanced,
    colback  = backcolour,
    fonttitle=\bfseries\rmfamily,
    fontupper=\scriptsize,
    title={\footnotesize\strut{#2}},
    #1
    }

\newtcolorbox{onebox}[2][]{
    enhanced, 
    center title,
    left*=0pt, right*=0pt,
    boxsep=2pt, left=5pt, right=5pt,
    skin first=enhanced,
    skin middle=enhanced,
    skin last=enhanced,
    colframe = mybluetitle!90,
  colback  = mybluetitle!10,
    fonttitle=\bfseries\rmfamily\fontfamily{phv}\selectfont,
    title={\footnotesize\strut{#2}  \refstepcounter{subsubsection} \addcontentsline{toc}{subsubsection}{\string\numberline{\thesubsubsection}#2}
    },
    #1
    }

\definecolor{logopurple}{HTML}{7E308E} 

\title{{Goal2Skill}: Long-Horizon Manipulation with Adaptive Planning and Progress Verification}

\author[1,2]{Zhen Liu}
\author[2]{Zhe Hu}
\author[1,2]{Xinyu Ning}
\author[2]{Hanlin Wang}
\author[1]{XinXin Xie}
\author[1]{Weize Li}
\author[2]{Zhipeng Tang}
\author[2]{Chongyu Wang}
\author[2]{Zejun Yang}
\author[1]{Yitong Liu}

\author[2,3, \dagger]{Zhongzhu Pu}

\affiliation[1]{Beijing University of Posts and Telecommunications}
\affiliation[2]{InspireOmni AI}
\affiliation[3]{Tsinghua University}

\abstract{

Robotic foundation models have advanced manipulation, yet long-horizon tasks remain challenging under partial observability and uncertain execution. Reliable task completion requires retaining relevant information, verifying intermediate outcomes, and adapting plans to execution feedback. We introduce \textbf{Goal2Skill}, a dual-system framework that couples adaptive planning, proactive memory, and a self-evolving verifier. A high-level VLM constructs and revises sub-task plans, while a low-level VLA executes action chunks. Periodic verification connects these processes by using current observations and task memory to determine when to continue, advance, or trigger reflective correction. Memory preserves information needed by upcoming steps and records previous attempts, supporting targeted revisions while retaining valid progress. The verifier combines lightweight models, rule-based checks, and VLM reasoning, and refines its assessment procedures through execution feedback, with candidate updates validated before adoption. Experiments on RoboMME and RMBench show that Goal2Skill achieves the highest average success rate among non-oracle methods, with particularly strong gains on tasks requiring persistent memory and multi-stage coordination.

}

\metadata[GitHub]{%
  \href{https://github.com/InspireOmniRobot/Goal2Skill}{%
    \textcolor{metablue}{\textnormal{InspireOmniRobot/Goal2Skill}}}}

\metadata[Correspondence]{%
  \href{mailto:puzz@inspireomni.ai}{%
    \textcolor{metablue}{\textnormal{puzz@inspireomni.ai}}}}

\begin{document}

\maketitle


\section{Introduction}

Robotic foundation models advance the range of manipulation tasks that robots can perform from natural-language instructions. Vision-language-action (VLA) models connect visual and linguistic representations to motor control, transferring semantic knowledge from pretrained models and manipulation skills from robot demonstrations~\citep{kawaharazuka2025vision, ma2024survey,black2024pi0,kim2024openvla}. More recently, world-action models (WAMs) and video-based policies couple action generation with predictions of future observations, leveraging spatiotemporal priors to capture physical dynamics and support anticipatory control~\citep{ye2026world,kim2026cosmos}. These advances improve the generalization of learned behavior. Nevertheless, \textit{long-horizon tasks} remain challenging because they require coordinating interdependent operations: retaining relevant information across interactions, verifying which outcomes have been achieved, and adapting subsequent decisions when execution changes the assumptions underlying an earlier plan.

\begin{figure}[!t]
    \centering
    \includegraphics[width=0.95\textwidth]{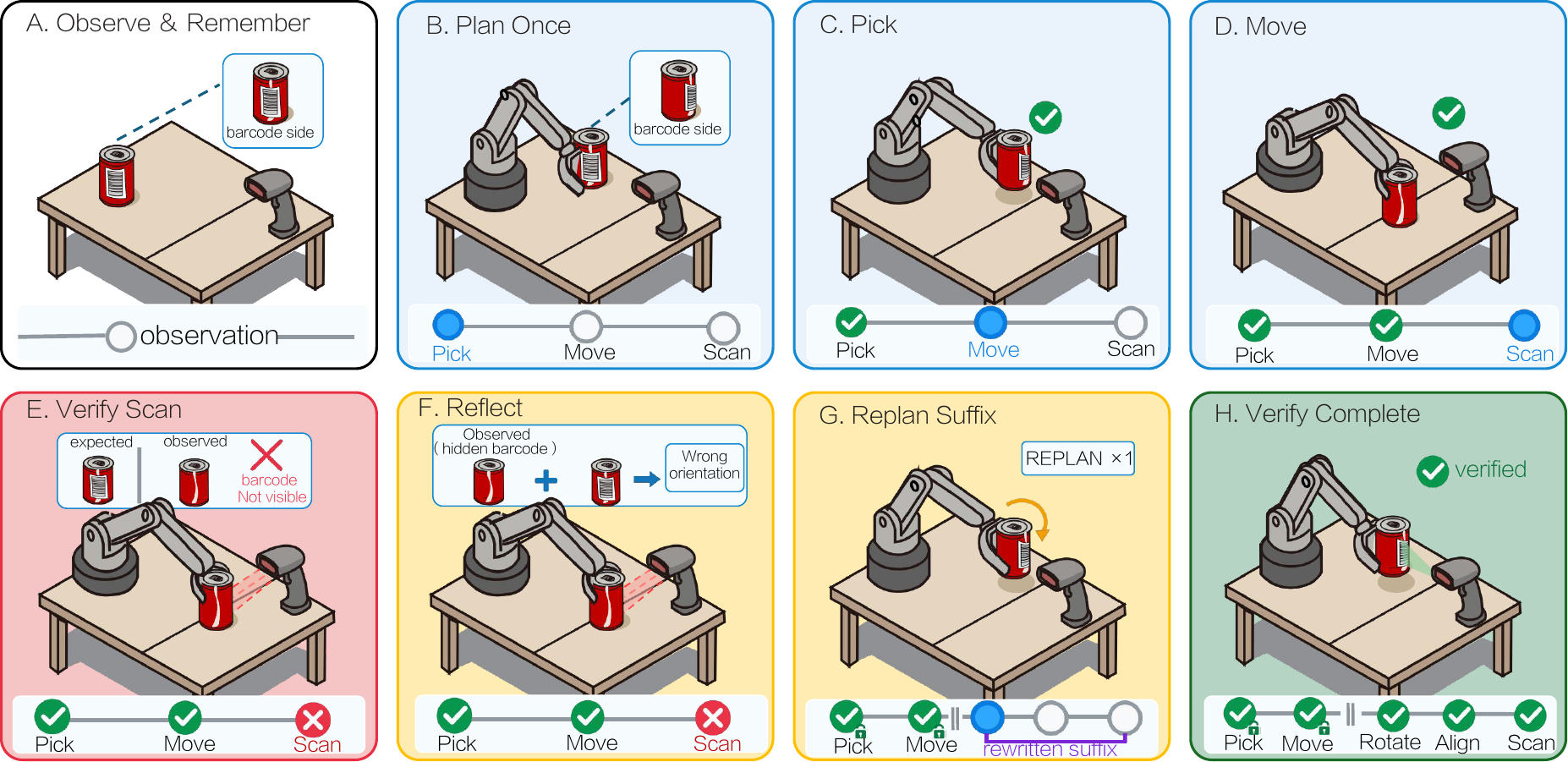}
    \caption{A long-horizon manipulation example.  The robot moves a Coke can to a scanner, but the first scan fails because the barcode is not visible. Reflection diagnoses a wrong orientation, and the remaining plan is rewritten to rotate and align the can before rescanning.
    }
    \vspace{-4mm}
    \label{fig:intro_example}
\end{figure}

Long-horizon manipulation requires coordinating planning, memory, and verification under partial observability and uncertain execution~\citep{zhou2023generalizable,chen2026rmbench,li2023behavior,li2026roboclaw,han2026robocerebra}. An initial plan may need revision as conditions change, information required by later steps may become unavailable, and executed actions may fail during execution. 
As shown in Figure~\ref{fig:intro_example}, a robot asked to scan a Coke can must first remember which side carries the barcode. After the can is moved to the scanner, verification reveals that the scan has failed because the barcode is not visible. Reflection attributes the failure to a wrong orientation rather than a failed grasp, so the system keeps the completed pick and move steps and rewrites only the remaining plan to rotate, align, and rescan the can. Such adaptations must combine current evidence with task history to revise the remaining plan while preserving valid progress. This illustrates the central challenge: deciding whether to continue, advance to the next sub-task, or revise the plan.

Recent dual-system architectures and related embodied frameworks offer a promising foundation by combining high-level reasoning with low-level execution, where a high-level model interprets goals and guides execution through plans, language subgoals, or skill invocations, while a low-level policy or controller realizes these decisions through physical interaction~\citep{team2025gemini, li2026roboclaw, zhang2026harness, chen2026show}. 
However, separating planning from execution does not ensure reliable adaptation as task conditions change. The system must retain context for future decisions and verify execution outcomes to determine when and how to revise the plan. Crucially, the evidence needed to establish success varies across tasks and scenes, motivating a \emph{self-evolving verifier} that refines its assessment criteria and procedures through validated execution feedback. This motivates coupling planning and proactive memory with evolving execution verification, so that improved progress assessment supports more reliable plan advancement and targeted correction.

We introduce \textbf{Goal2Skill}, a dual-system framework that couples \emph{adaptive planning}, \emph{proactive memory}, and \emph{execution verification} for long-horizon manipulation. A high-level VLM brain decomposes the goal into sub-tasks, while a low-level VLA executes them through continuous visuomotor control. Periodic verification assesses task progress and determines whether to continue execution, advance to the next sub-task, or trigger reflection for local correction and replanning. Task memory supports these decisions by preserving information needed by upcoming steps and updating task context from execution outcomes. Together, these mechanisms enable the brain to adapt its plan while retaining valid progress and accounting for previous attempts.

Central to this coordination is a \emph{self-evolving verifier} that governs task progression by determining when to continue execution, advance to the next step, or trigger correction. It combines lightweight tools, rule-based checks, and VLM reasoning to balance reliability and computational cost. Execution feedback informs iterative refinements to its assessment criteria and checking procedures, with candidate updates validated before adoption. Operating between frequent VLA execution and sparse planning, the verifier provides timely feedback for reflection and plan revision while adapting progress assessment to different tasks and scenes.

We evaluate Goal2Skill on RoboMME and RMBench tasks spanning diverse manipulation settings. It achieves the highest average success rate with particularly strong gains on tasks requiring persistent memory and multi-stage coordination. Qualitative analyses illustrate how execution feedback refines verification procedures and how reflection guides replanning when execution invalidates the initial plan. Our contributions are threefold:
\begin{itemize}[nolistsep,leftmargin=1.4em]
    \item We introduce an adaptive planning mechanism that combines initial task decomposition with evidence-guided retries, local adjustments, and revision of the remaining plan.
    \item We develop proactive task memory with a self-evolving verifier  to guide task progression and refine verification through execution feedback.
    \item Experiments on RoboMME and RMBench show that Goal2Skill achieves the highest average success rate  with strong gains on tasks requiring persistent memory and multi-stage coordination.
\end{itemize}

\section{Related Work}

\noindent\textbf{Long-Horizon VLA and Agentic Manipulation.}
VLA models unify visual perception, language understanding, and motor control. Systems such as RT-2, OpenVLA, $\pi_0$, and RDT-1B demonstrate instruction-following and generalization capabilities through multimodal and robot-data pretraining~\citep{rt22023arxiv,kim2024openvla,black2024pi0,liu2024rdt}. World-action models (WAMs) such as Cosmos Policy couple action generation with predicted future observations and values to support anticipatory visuomotor control~\citep{ye2026world,kim2026cosmos}.
Recent work also tackles long-horizon control. For example, Long-VLA introduces phase-aware modeling to improve multi-step manipulation~\citep{fan2025long}. 

Complementary approaches coordinate high-level reasoning with low-level execution~\citep{chen2026show,ding2026zetta}. Gemini Robotics and RoboClaw combine embodied reasoning with manipulation policies~\citep{team2025gemini,li2026roboclaw}; LoHo-Manip updates progress-aware remaining plans~\citep{liu2026long}; and Harness VLA composes VLA behaviors and analytic primitives through memory-guided orchestration~\citep{zhang2026harness}. Our work focuses on memory-grounded task progression, coupling adaptive planning and proactive memory with a self-evolving verifier that uses execution evidence to guide plan advancement and targeted recovery.


\noindent\textbf{Memory, Execution Verification, and Reflective Recovery.}
Memory-augmented policies retain information beyond the current observation. MemoryVLA combines perceptual and semantic memory~\citep{shi2025memoryvla}, MEM integrates both short-term and long-term contexts~\citep{torne2026mem}, and ReMem-VLA carries recurrent representations across observation windows~\citep{li2026remem}. RMBench further highlights the challenges of memory-dependent manipulation and the limitations of simply extending observation history~\citep{chen2026rmbench}. 

Recent work also uses execution feedback to assess progress and support recovery. ProgressVLA guides action generation with learned progress estimates~\citep{yan2026progressvla}; Phoenix and PhysReflect-VLA connect reflection with motion correction and physical consistency assessment~\citep{xia2025phoenix,yang2026physreflect}; and CheckVLA detects deviations through action-conditioned world-model predictions~\citep{liu2026checkvla}. Goal2Skill couples these concerns at the task level: proactive memory preserves evidence needed by upcoming steps, while verification guides plan advancement and reflective correction. Its verifier further refines its checking criteria and procedures through validated execution feedback, supporting adaptation across tasks and scenes.

\section{System Architecture}
\label{sec:architecture}


We study long-horizon robotic manipulation specified by a natural-language goal $\mathcal{G}$. At each environment step $t$, the robot receives an observation $o_t$ comprising visual input and proprioceptive state, and executes a continuous motor action $a_t$. The objective is to maximize the probability of achieving $\mathcal{G}$ within a global execution budget of $T$ steps.

The environment is partially observable: information needed for a decision may be available only in earlier observations or interaction outcomes. 
Long-horizon execution therefore requires combining current observations with retained context to coordinate interdependent stages and adapt subsequent decisions to observed outcomes.

\begin{figure*}[t]
    \centering
    \includegraphics[width=0.95\textwidth]{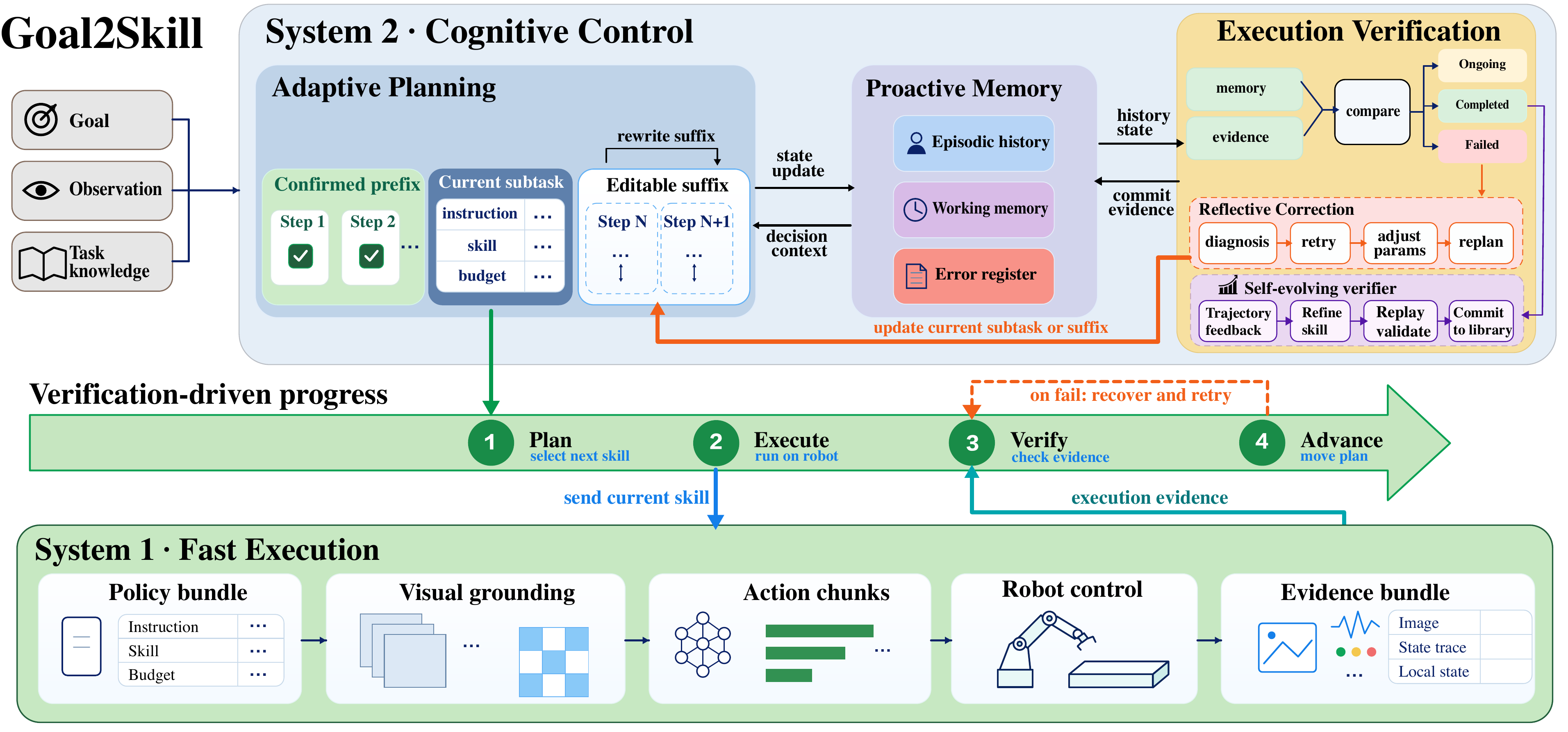}
    \vspace{-2mm}
    \caption{Overview of Goal2Skill. The high-level system integrates adaptive planning, proactive memory, progress verification, and reflective recovery. The low-level executor performs task-conditioned visual grounding and generates action chunks, returning execution evidence for verification and memory updates. The verifier guides task progression by determining when to continue execution, advance to the next sub-task, or trigger corrective reasoning. This closed loop supports targeted plan correction during long-horizon manipulation.
}
    \vspace{-2mm}
    \label{fig:architecture}
\end{figure*}

\subsection{Framework Overview}
\label{sec:overall}

As illustrated in Figure~\ref{fig:architecture}, Goal2Skill adopts a dual-system architecture comprising a high-level VLM reasoning system and a low-level VLA executor. The \textbf{high-level system} coordinates adaptive planning, proactive memory, execution verification, and reflective recovery. The \textbf{low-level executor} carries out the active sub-task through visual grounding and successive action chunks conditioned on updated observations.

A \textbf{memory-grounded verifier} links low-level execution to high-level task progression. At periodic checkpoints, it assesses progress on the active sub-task: verified completion advances execution to the next sub-task; detected failures or an exhausted sub-task budget trigger reflection. A reflection module diagnoses the discrepancy and informs a retry, a local adjustment, or a revision of the remaining plan. After the final planned sub-task, the verifier checks whether the global goal has been achieved and, if necessary, initiates further recovery within the remaining execution budget. Throughout this process, memory preserves information needed by future steps and is updated with verified outcomes and previous recovery attempts.

The framework separates three temporal scales: frequent action-chunk execution, periodic verification, and sparse planning or replanning, with reflection triggered by execution discrepancies. To balance responsiveness and computational cost, the verifier uses lightweight models or rule-based checks when sufficient and invokes VLM reasoning when needed. Through validated execution feedback, the verifier iteratively refines its skills, enabling a self-evolving verification mechanism that adapts to new tasks and supports more generalizable task progression.



\subsection{High-Level Planning and Coordination}
\label{sec:high_level_planner}

The high-level system leverages a pretrained VLM $\Phi$ as brain and coordinates adaptive planning (\S~\ref{sec:planning}), proactive memory management (\S~\ref{sec:proactive_memory}), progress verification (\S~\ref{sec:progress_verification}), and reflective correction (\S~\ref{sec:reflective_correction}). These components share structured task memory, allowing decisions to account for the global goal, observed progress, and previous attempts. Planning occurs initially and when revision is needed, verification runs periodically during execution, and reflection is triggered when corrective intervention is required.

\subsubsection{Task Decomposition and Adaptive Planning}
\label{sec:planning}

Given a natural-language goal $\mathcal{G}$ and an initial observation $o_0$, the planner generates an ordered sequence of sub-tasks:
\begin{equation}
\mathcal{P}
= \langle \tau_1,\tau_2,\ldots,\tau_K\rangle
= \Phi_{\mathrm{plan}}(\mathcal{G},o_0).
\label{eq}
\end{equation}
Each sub-task is represented as 
$
\tau_k =
(\ell_k,\delta_k,j_k),
\label{eq}
$
where $\ell_k$ is its language instruction, $\delta_k$ sets the execution budget per attempt, and $j_k$ selects a low-level manipulation skill.

The initial plan captures dependencies between task stages and guides subsequent execution. Verified sub-task completion advances to the next sub-task, provided its prerequisites are satisfied, without regenerating the plan. When execution of the sub-task deviates from expectations, verification and reflection guide corrections at three scopes: retrying the active sub-task, adjusting its local specification, or asking the planner to revise the remaining sub-task sequence. This allows the system to address local failures while retaining parts of the plan that remain valid.

\subsubsection{Proactive Memory Management}
\label{sec:proactive_memory}


Long-horizon manipulation tasks require retaining history contexts, information that may disappear from view, together with the outcomes of earlier interactions. We maintain an episode-level memory:
\begin{equation}
\mathcal{M}_t =
(\mathcal{H}_t,\mathcal{W}_t,\mathcal{E}_t),
\label{eq}
\end{equation}
where the three components serve distinct roles.

\textbf{Episodic history $\mathcal{H}_t$} maintains a compact, chronological account of task progress across sub-tasks. It summarizes previous interactions, observed changes in robot and world state, and verified outcomes. Information that remains relevant to later stages, such as a previously observed target ordering, is retained even after it leaves the current view.

\textbf{Working memory $\mathcal{W}_t$} tracks the active sub-task at $t$-th step. It records the state observed at the start of the current sub-task, recent execution evidence, and progress toward the expected outcome of this step. Comparing the current state with this local reference helps the verifier determine what has changed and whether the intended sub-task execution has succeeded.

\textbf{The error register $\mathcal{E}_t$} records failures and recovery attempts throughout the episode. Each entry includes the observed discrepancy, hypothesized cause, proposed correction, and its outcome when available. These records support reflection on both the current failure and similar previous attempts.

Memory updates follow the execution cycle. When a sub-task or a new attempt begins, working memory is initialized from the current observation and relevant episodic context. At each verification checkpoint, recent evidence and the verification result update its local progress record. When a sub-task completes, or an attempt ends in failure, its relevant information is summarized into episodic history before working memory is refreshed for the next attempt or sub-task. Recovery decisions are recorded in the error register and updated when their outcomes become available:
\begin{equation}
\mathcal{M}_t^{+}
=
\Phi_{\mathrm{mem}}
\left(
\mathcal{M}_t,\mathcal{P}_{t:T},
\mathcal{Z}_t,v_t
\right),
\label{eq}
\end{equation}
where $\mathcal{M}_t^{+}$ denotes memory after the update, $\mathcal{P}_{t:T}$ is the remaining plan, $\mathcal{Z}_t$ contains recent observations and executed actions, and $v_t$ is the verification output defined below.
Importantly, the remaining plan makes memory maintenance proactive: it identifies information required by subsequent steps, guiding what should be retained when observed. 

To bound context length, recent execution evidence is retained in a limited buffer, while older interactions are compressed into episodic summaries. Compression preserves information relevant to upcoming steps, unresolved discrepancies, and previous recoveries. This keeps working memory focused on the active sub-task without losing the global context needed for planning and reflection.

\subsubsection{Memory-Grounded Progress Verification}
\label{sec:progress_verification}

The verifier assesses the active sub-task after every $n$ executed
action chunks, where $n$ controls the verification interval.
At the start of a sub-task, it retrieves a task-relevant verification skill from the verification skill library using the sub-task specification, current scene, and task memory.
Each verification skill specifies its trigger condition, required evidence, checking procedure, and decision criterion.
Each time, the verifier produces
\begin{align}
& v_t=(e_t, c_t)
=V_{\theta_k}(\mathcal{Z}_t,\tau_k,\mathcal{M}_t), \\
&  c_t\in
\{\textsc{completed},\textsc{ongoing},\textsc{failed}\},
\label{eq}
\end{align}
where $\theta_k$ denotes the verification rule and $c_t$ is the assessed status with reason $e_t$. 
A sub-task is \textsc{completed} when the evidence supports its expected outcome. It is \textsc{failed} when the evidence establishes a discrepancy requiring corrective intervention. Otherwise, it returns \textsc{ongoing} when the execution is still in progress or the evidence is insufficient to confirm completion.

To balance reliability and efficiency, the verification combines rule-based checks, lightweight models, and VLM. Inexpensive checks handle directly measurable conditions, while ambiguous, history-dependent, or situationally complex judgments invoke VLM reasoning. Where necessary, evidence across multiple checkpoints is used to distinguish persistent failures from transient observations.
A \textsc{completed} assessment advances the existing plan, while \textsc{ongoing} retains the active sub-task. A \textsc{failed} assessment triggers reflection. After the final planned sub-task, the verifier also checks the global goal; completing the planned sequence alone is insufficient to declare task success.

Notably, we introduce a \emph{self-evolving verifier mechanism} that improves its verification procedures from prior task trajectories and environment feedback.
The verifier retrieves task-relevant verification skills during online execution, while accumulated trajectories are used to identify gaps and update existing skills or introduce new ones. The updated skills are validated through trajectory replay before being committed to the verification skill library. This allows the verifier to adapt its progress assessment from execution experience.

\subsubsection{Reflective Correction and Replanning}
\label{sec:reflective_correction}

Reflection is triggered by a detected execution failure, an exhausted sub-task budget, or an unmet global goal after the planned sequence ends. The reflection module examines the discrepancy between the intended outcome and observed execution:
\begin{equation}
(d_t,\rho_t,\tau'_t)
=
\Phi_{\mathrm{reflect}}(\mathcal{F}_t,\mathcal{M}_t),
\label{eq}
\end{equation}
where $\mathcal{F}_t$ contains the relevant sub-task evidence. The output comprises a diagnosis $d_t$, a recovery decision $\rho_t$, and an optional proposed modification $\tau'_t$. The diagnosis distinguishes observed discrepancies from hypothesized causes and explains the evidence supporting the proposed correction.

The recovery decision selects among three scopes. \textsc{Retry} repeats the active sub-task when another attempt is justified without changing its specification. \textsc{Adjust} revises the active sub-task's instruction, grounding, or supported execution parameters ($\tau'_t$) while retaining the remaining sequence. \textsc{Replan} invokes the planner with the global goal to revise unfinished sub-tasks.
Recovery generally escalates from a justified retry to local adjustment and then to broader replanning when previous corrections fail. It can proceed directly to replanning when evidence shows that the remaining sequence is no longer appropriate.

Diagnoses and proposed corrections are recorded in $\mathcal{E}_t$ when issued, then supplemented with their observed outcomes. These records help subsequent reflection avoid repeating ineffective recovery actions. Recovery actions, including retries, local adjustments, and replanning, are bounded, and all execution remains subject to the global task budget.

\subsection{Low-Level Executor}
\label{sec:low_level_executor}

The low-level executor carries out the current sub-task through closed-loop visuomotor control. At each action-generation step, it receives an observation $o_t=(I_t,s_t)$, consisting of RGB images $I_t$ and proprioceptive state $s_t$, together with the sub-task instruction $\ell_k$. The skill index $j_k$ selects a manipulation policy, which generates a chunk of continuous actions.

\subsubsection{Task-Conditioned Visual Filtering}
\label{sec:masked_perception}

In cluttered scenes, visually salient but task-irrelevant objects can distract the low-level policy from the target object. Following~\citet{liu2026hsc}, the executor suppresses such regions to focus visuomotor control on the relevant objects. When a sub-task begins, a spatial grounding step uses the current image, sub-task instruction, and relevant task context to identify distractor bounding boxes $\mathcal{B}_{k,t}$. 
A segmentation model $\mathcal{S}$ converts these boxes into a binary mask $Q_t$, where one denotes a distractor pixel. A lightweight tracking module then propagates the mask across subsequent observations. The filtered image is
$
\hat{I}_t = I_t \odot (1-Q_t),
\label{eq}
$
where $\odot$ denotes element-wise multiplication. This operation preserves the coordinates and pixel values of unmasked regions. Grounding is refreshed when the sub-task changes, a corrective adjustment modifies the relevant regions, or mask tracking becomes unreliable.
Notably, only the action policy receives the filtered image. Original observations remain available to high-level modules, preserving access to evidence outside the executor's immediate task focus.

\subsubsection{Action Generation}
\label{sec:skill_library}

The executor exposes a library of manipulation policies $\Pi=\{\pi_1,\ldots,\pi_J\}$ through a common execution interface. The skill index $j_k$ selects the policy for the active sub-task. Conditioned on the filtered image, proprioceptive state, and instruction, the policy model generates an action chunk:
$
\mathbf{A}_t=(a_t,\ldots,a_{t+H-1}),
\label{eq}
$
where $H$ is the prediction horizon. This interface allows task-specific policies (e.g., VLA models) to be invoked through the same sub-task representation.

Execution follows a receding-horizon procedure: the policy model generates an action chunk, obtains an updated observation, and generates the next chunk. The active sub-task is retained while its verification status remains \textsc{ongoing}. Verified completion ends the sub-task, while a corrective intervention or exhaustion of the per-attempt budget $\delta_k$ interrupts execution and returns control to the high-level coordination system. Execution observations and actions are recorded for periodic verification and memory updates. This separation allows the policy to respond to local scene changes while reserving high-level planning for task decomposition and necessary revisions.

\section{Experimental Setup}
\label{sec:exp_setup}

We evaluate whether Goal2Skill improves long-horizon manipulation by maintaining task-relevant information, verifying execution progress, and adapting subsequent decisions from execution feedback. Experiments are conducted on two memory-dependent manipulation benchmarks: RoboMME~\citep{dai2026robomme} and RMBench~\citep{chen2026rmbench}.

\noindent\textbf{RoboMME.}
RoboMME benchmark comprises 16 tasks across four suites: \emph{Counting}, \emph{Permanence}, \emph{Reference}, and \emph{Imitation}. These suites assess progress tracking, object permanence, referential reasoning, and reproduction of previously observed behaviors.
We evaluate all 16 tasks and highlight six representative tasks in the main paper. \emph{PickXtimes} requires tracking repeated interactions, while \emph{ButtonUnmaskSwap} requires retaining state changes from earlier interactions. \emph{VideoRepick} tests recall of previously observed targets, and \emph{VideoPlaceOrder} additionally requires preserving their temporal order. \emph{PickHighlight} requires incorporating information revealed during interaction, while \emph{VideoPlaceButton} requires verifying an intermediate manipulation before proceeding. Following the benchmark protocol, we use 100 training demonstrations and 50 test episodes per task, yielding 800 test episodes across the full benchmark.
Complete results are provided in Appendix~\ref{app:robomme_results}.

\noindent\textbf{RMBench.}
We include five representative tasks: two M(1) tasks, \emph{Observe and Pick Up} and \emph{Rearrange Blocks}, and three M(n) tasks, \emph{Battery Try}, \emph{Blocks Ranking Try}, and \emph{Press Button}. The M(1) tasks emphasize relatively local memory dependencies, whereas the M(n) tasks require retaining context across multiple stages. 
We report per-task success rates and the unweighted mean.

\noindent\textbf{Training Details.}
For RoboMME, we follow the original setting. The low-level policy is initialized from the $\pi_{0.5}$ with an action horizon of 20. For RMBench, we use the official $\pi_{0.5}$ fine-tuning configuration. Demonstrations are segmented into sub-tasks to provide supervision for the hierarchical execution pipeline. More implementation details are provided in Appendix~\ref{sec:app_training_details}.

\noindent\textbf{Baselines.}
For RoboMME, we compare against $\pi_{0.5}$ with and without past-action context, historical-frame sampling via MME-VLA (FrameSamp)~\citep{dai2026robomme}, keyframe retrieval via MemER~\citep{sridhar2025memer}, and visual memory via SAM2Act+~\citep{fang2025sam2act}. We also include GroundSG+QwenVL, which predicts subgoals from current observations and subgoal history, and GroundSG+Oracle, which uses simulator-provided ground-truth subgoals as an oracle reference~\citep{dai2026robomme}. 
For RMBench, we include Diffusion Policy (DP)~\citep{chi2025diffusion}, ACT~\citep{zhao2023learning},  $\pi_{0.5}$~\citep{intelligence2025pi_} and X-VLA~\citep{zheng2025x}. We additionally consider
 memory-augmented methods including MemER~\citep{sridhar2025memer}.



\section{Results and Analysis}
\subsection{Main Results on RoboMME}
\label{sec:quantitative}

As in Table~\ref{tab:robomme_main}, Goal2Skill reaches 70.0\% on the six representative tasks, outperforming the strongest non-oracle baseline by over 25 percentage points. Goal2Skill ranks first on four of these tasks and second on the remaining two. Moreover, Goal2Skill achieves a higher average success rate across all 16 RoboMME tasks than all other baselines (Table~\ref{tab:robomme_full}), demonstrating its strong performance.

\begin{table}[t]
    \fontsize{10}{12}\selectfont
    \centering
    \caption{
    Success rates (\%) on RoboMME tasks.
    Average denotes the average over the six selected tasks.
    Best and second-best results among non-oracle methods are shown in
    \textbf{bold} and \underline{underline}, respectively.
    Complete results on all 16 tasks are reported in
    Appendix Table~\ref{tab:robomme_full}. 
    All RoboMME baselines are reproduced using the officially released checkpoints.
    }
    \vspace{-4mm}
    \label{tab:robomme_main}
    \vspace{0.5ex}
    \setlength{\tabcolsep}{3.5pt}
    \renewcommand{\arraystretch}{0.95}
    \resizebox{\textwidth}{!}{%
    \begin{tabular}{l*{6}{c}>{\columncolor{blue!10}}c}
        \toprule
        Method
        & \shortstack[c]{Pick\\Xtimes}
        & \shortstack[c]{Button\\UnmaskSwap}
        & \shortstack[c]{Video\\Repick}
        & \shortstack[c]{Video\\PlaceOrder}
        & \shortstack[c]{Pick\\Highlight}
        & \shortstack[c]{Video\\PlaceButton}
        & \cellcolor{blue!10}\textbf{Average}
        \\
        \midrule
        
        \textcolor{GroupText}{\small\itshape $\bullet$\ Base Policies:} \\
        $\pi_{0.5}$
        & 42.0
        & 6.0
        & 0.0
        & 26.0
        & 12.0
        & 32.0
        & 19.7
        \\
        $\pi_{0.5}$ w/ past actions
        & 58.0
        & 16.0
        & 8.0
        & 18.0
        & 12.0
        & 24.0
        & 22.7\\
        \midrule
        \multicolumn{8}{l}{
            \textcolor{GroupText}{\small\itshape $\bullet$\ Memory-Augmented VLA Policies:}
        } \\
        MME-VLA (FrameSamp)
        & 82.0
        & 12.0
        & 32.0
        & 28.0
        & 24.0
        & \textbf{54.0}
        & 38.7\\
        MemER
        & 78.0
        & 24.0
        & 32.0
        & \underline{36.0}
        & \textbf{82.0}
        & 16.0
        & \underline{44.7}
        \\
        SAM2Act+
        & 76.0
        & \underline{26.0}
        & 6.0
        & 20.0
        & 18.0
        & 24.0
        & 28.3
        \\
        \midrule

        \multicolumn{8}{l}{
            \textcolor{GroupText}{\small\itshape $\bullet$\ Dual Systems:}
        } \\

        GroundSG+QwenVL
        & \underline{90.0}
        & 10.0
        & \underline{34.0}
        & \underline{36.0}
        & 24.0
        & 48.0
        & 40.3
        \\

        \textbf{Goal2Skill}
        & \textbf{92.0}
        & \textbf{50.0}
        & \textbf{60.0}
        & \textbf{90.0}
        & \underline{76.0}
        & \underline{52.0}
        & \textbf{70.0}
        \\

        \midrule

        \multicolumn{8}{l}{
            \textcolor{GroupText}{\small\itshape $\bullet$\ Oracle Reference:}
        } \\

        GroundSG+Oracle
        & 100.0
        & 86.0
        & 98.0
        & 98.0
        & 52.0
        & 100.0
        & 89.0
        \\

        \bottomrule
    \end{tabular}%
    }
\vspace{-2mm}
\end{table}

The largest gains occur on \emph{ButtonUnmaskSwap}, \emph{VideoRepick}, and \emph{VideoPlaceOrder}, which require retaining earlier state changes, recalling previously observed targets, and preserving temporal order, respectively. These results are consistent with its emphasis on preserving task-relevant context and using execution feedback to guide subsequent decisions.

\begin{wrapfigure}{r}{0.45\textwidth}
\vspace{-4mm}
    \centering
    \includegraphics[width=\linewidth]{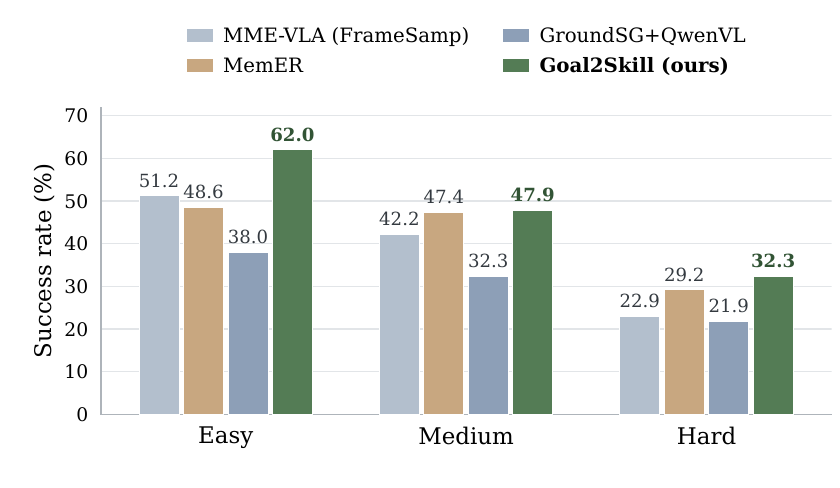}
    \vspace{-6mm}
    \caption{Success rates by difficulty, averaged over all 16 RoboMME tasks.}
    \label{fig:difficulty_robomme}
    \vspace{-6mm}
\end{wrapfigure}

Comparisons with baselines further clarify this pattern. Adding past actions to $\pi_{0.5}$ improves its six-task average only from 19.7\% to 22.7\%, suggesting that this form of temporal context alone is insufficient. GroundSG+QwenVL nearly matches Goal2Skill on \emph{PickXtimes}, but remains substantially behind on the three tasks above. Although MemER performs better on \emph{PickHighlight} and FrameSamp slightly outperforms Goal2Skill on \emph{VideoPlaceButton}, Goal2Skill remains competitive on both while providing larger gains on tasks with extended state and ordering dependencies.

We further show that the overall advantage spans all three benchmark-defined difficulty levels in Figure~\ref{fig:difficulty_robomme}. Across all 16 tasks, Goal2Skill achieves the best results when averaged over all episodes,
yielding the highest non-oracle average at each level. 
Specifically, it retains a substantial advantage on Hard \emph{VideoPlaceOrder}, and leads on
\emph{ButtonUnmaskSwap} at all levels. These findings highlight its effectiveness in carrying historical context into later execution stages, including under more demanding conditions. Detailed results are provided in Appendix Tables~\ref{tab:robomme_difficulty_cp}
and~\ref{tab:robomme_difficulty_ri}.


\subsection{Main Results on RMBench}

Table~\ref{tab:rmbench_detailed} shows that Goal2Skill achieves the highest success rate across the five selected RMBench tasks and the highest success rate on four of the five tasks, supporting its effectiveness beyond RoboMME.
The advantage is particularly pronounced on tasks with multi-stage memory dependencies. Goal2Skill reaches 23.0\% on the M(1) tasks, compared with 15.0\% for the strongest baseline, and 38.7\% on the M(n) tasks, compared with 9.0\%. On \emph{Battery Try}, which requires using outcomes from previous attempts to guide subsequent actions, Goal2Skill significantly outperforms the baselines. On \emph{Blocks Ranking Try}, which requires maintaining ordering information across interactions, it reaches 60\%, compared with 10\% for the strongest baseline. These gains are consistent with the framework design on tracking intermediate outcomes and reusing task-relevant information in subsequent decisions.

Goal2Skill also improves \emph{Rearrange Blocks} success rate, and achieves 10\% on \emph{Press Button}, where all compared baselines obtain zero success. On \emph{Observe and Pick Up}, it remains close to the best baseline. Together with the RoboMME results, these findings indicate that Goal2Skill's main advantage lies in coordinating tasks that require information and execution outcomes to remain available across multiple stages. Ablations in Appendix~\ref{sec:exp_ablation} further show the effectiveness of each module.

\begin{table}[t]
\fontsize{10}{12}\selectfont
    \centering
    \caption{
    Detailed results on five RMBench tasks.
    We report success rates (\%) together with the corresponding group-wise and overall averages. 
    Baseline results are taken from ~\cite{yang2026eventvla}. 
    To better isolate the contribution of high-level reasoning and memory, we only compare with methods that use standard low-level policies without specific architectural modifications.
    }
    \label{tab:rmbench_detailed}

    \vspace{-3mm}
    \small
    \setlength{\tabcolsep}{3.7pt}
    \renewcommand{\arraystretch}{0.9}
    \resizebox{\textwidth}{!}{%
    \begin{tabular}{lccccccc>{\columncolor{matchagreen}}c}
        \toprule
        \multirow{2}{*}{Method}
        & \multicolumn{2}{c}{M(1) Tasks}
        & \multicolumn{3}{c}{M(n) Tasks}
        & \multicolumn{3}{c}{Average}
        \\
        \cmidrule(lr){2-3}
        \cmidrule(lr){4-6}
        \cmidrule(lr){7-9}

        &
        \shortstack[c]{Observe and\\Pick Up}
        &
        \shortstack[c]{Rearrange\\Blocks}
        &
        \shortstack[c]{Battery\\Try}
        &
        \shortstack[c]{Blocks Ranking\\Try}
        &
        \shortstack[c]{Press\\Button}
        &
        M(1)
        &
        M(n)
        &
        \cellcolor{matchagreen}\textbf{Total}
        \\
        \midrule

        \multicolumn{8}{l}{
            \textcolor{GroupText}{\small\itshape $\bullet$\ Base Policies:}
        }\\

        DP
        & 1.0
        & 0.0
        & 10.0
        & \underline{10.0}
        & \underline{0.0}
        & 0.5
        & 6.7
        & 4.2
        \\

        ACT
        & 1.0
        & \underline{29.0}
        & 19.0
        & 0.0
        & \underline{0.0}
        & \underline{15.0}
        & 6.3
        & 9.8
        \\



        $\pi_{0.5}$
        & \textbf{9.0}
        & 13.0
        & 16.0
        & 6.0
        & \underline{0.0}
        & 11.0
        & 7.3
        & 8.8
        \\

        X-VLA
        & \textbf{9.0}
        & 13.0
        & 26.0
        & 1.0
        & \underline{0.0}
        & 11.0
        & \underline{9.0}
        & 9.8
        \\

        \midrule
        \multicolumn{8}{l}{
            \textcolor{GroupText}{\small\itshape $\bullet$\ Memory-Augmented VLA Policies:}
        }\\

        MemER
        & 7.0
        & 17.0
        & \underline{27.0}
        & 0.0
        & \underline{0.0}
        & 12.0
        & \underline{9.0}
        & \underline{10.2}
        \\


        \midrule

        \multicolumn{8}{l}{
            \textcolor{GroupText}{\small\itshape $\bullet$\ Dual Systems:}
        }\\

        \textbf{Goal2Skill}
        & \underline{8.0}
        & \textbf{38.0}
        & \textbf{46.0}
        & \textbf{60.0}
        & \textbf{10.0}
        & \textbf{23.0}
        & \textbf{38.7}
        & \textbf{32.4}
        \\

        \bottomrule
    \end{tabular}%
    }
\vspace{-4mm}
\end{table}







\par

\subsection{Qualitative Analysis}
\label{sec:qualitative}

\begin{figure}[t]
    \centering
    \includegraphics[width=0.99\linewidth]{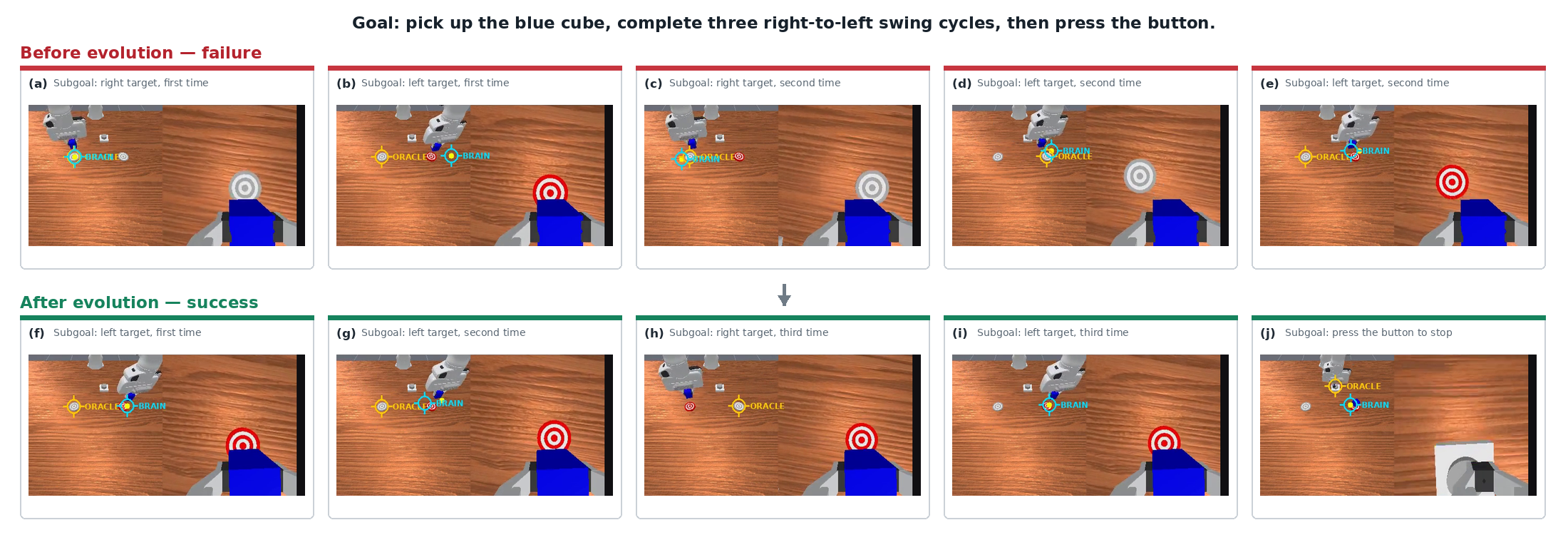}
    \vspace{-2mm}
    \caption{A case indicating how the verification-driven skill evolution improves performance for swing counting.}
    \vspace{-2mm}
    \label{fig:verify_skill}
\end{figure}

\noindent\textbf{Feedback-Driven Verifier Evolution.}
Figure~\ref{fig:verify_skill} illustrates how feedback improves progress assessment. The task requires three complete right-to-left swing cycles before pressing a button. The system initially infers progress from transitions in the planned subgoal sequence. These transitions do not reliably establish that the corresponding physical movements have been completed, causing the internal count to diverge from actual execution and potentially triggering an early button press.

To address this discrepancy, Goal2Skill refines the verifier with a dedicated counting procedure. It detects arrivals at the left and right target regions, suppresses repeated detections of the same arrival, and counts a cycle only after observing the required sequence of transitions. In the illustrated rollout, this procedure correctly tracks all three cycles before allowing progression to the button-press subgoal. The example demonstrates how execution failures can inform task-specific verification procedures that ground plan advancement in observable outcomes.

\begin{wrapfigure}{r}{0.49\textwidth}
\vspace{-2mm}
    \centering
    \includegraphics[width=\linewidth]{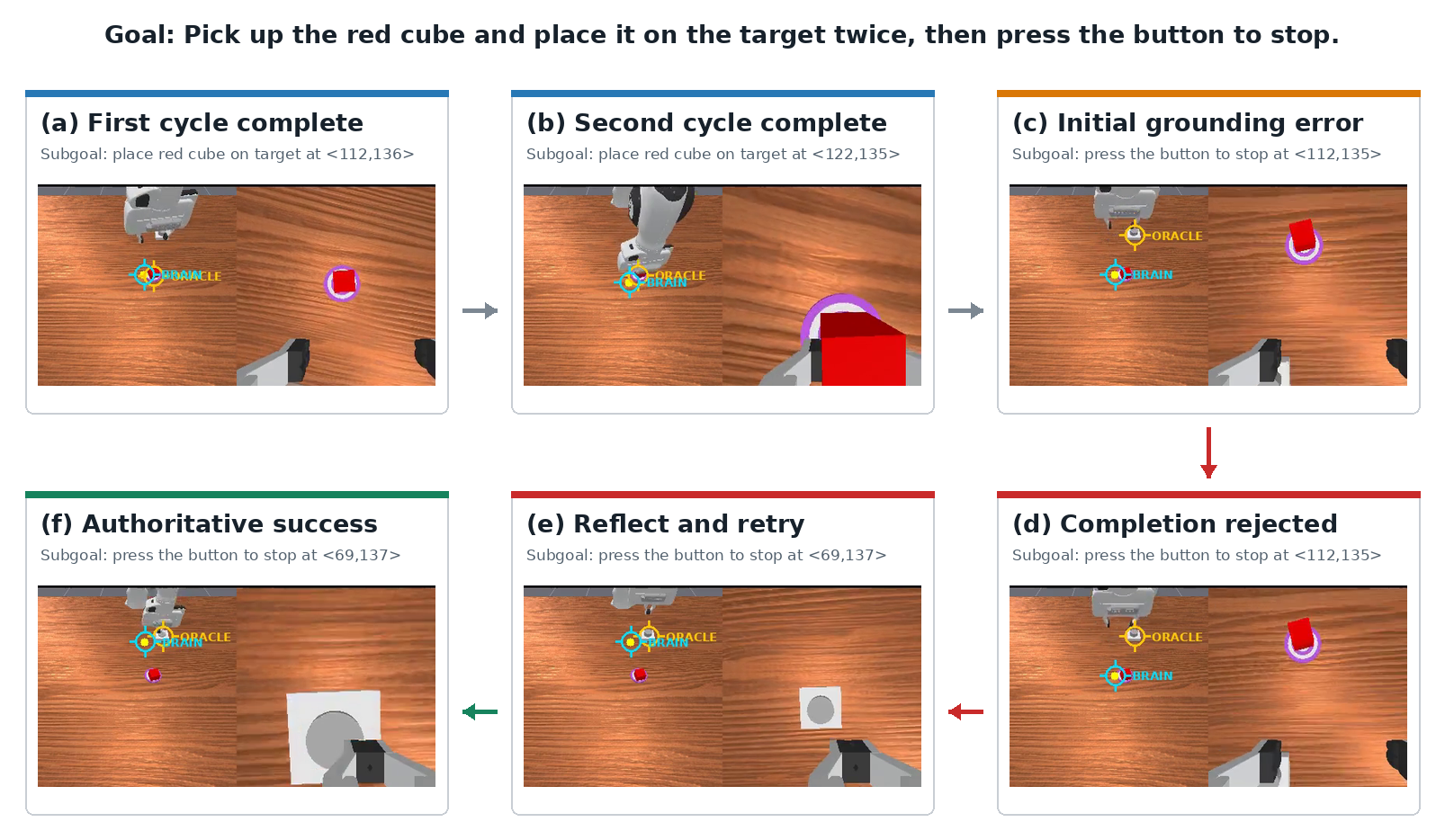}
    \vspace{-7mm}
    \caption{Reflection-based replanning after a verification failure.}
    \label{fig:reflection_case}
    \vspace{-4mm}
\end{wrapfigure}

\noindent\textbf{Reflection-Guided Replanning.}
Figure~\ref{fig:reflection_case} shows how Goal2Skill adapts when execution invalidates an assumption of the initial plan. Early sub-tasks proceed as expected, but a later verification step reveals that the current state does not satisfy the prerequisites for the next planned action.

The discrepancy triggers reflection over the updated observation, task memory, and execution history. This diagnosis informs a revision of the remaining plan, producing a sub-task appropriate to the current state while retaining previous progress. Execution then resumes from the revised subgoal.

This shows the complementary roles of verification and reflection: verification identifies the mismatch between expected and observed state, while reflection guides how the plan should change. The interaction enables replanning as execution unfolds. More samples are in Appendix~\ref{sec:app_qualitative}.

\section{Conclusion}

We presented Goal2Skill, a dual-system framework that couples adaptive planning, proactive memory, and memory-grounded verification for long-horizon manipulation. By connecting execution evidence to task progression and reflective correction, the framework supports targeted plan revisions while preserving valid progress. Feedback-driven verifier evolution further enables task-specific refinement of progress assessment. Experiments on RoboMME and RMBench demonstrate improved success, particularly when information must be retained and reused across multiple stages. Future work will explore self-improvement across more diverse tasks and embodiments.




\clearpage

\bibliography{main}

@article{torne2026mem,
  title={MEM: Multi-Scale Embodied Memory for Vision Language Action Models},
  author={Torne, Marcel and Pertsch, Karl and Walke, Homer and Vedder, Kyle and Nair, Suraj and Ichter, Brian and Ren, Allen Z and Wang, Haohuan and Tang, Jiaming and Stachowicz, Kyle and others},
  journal={arXiv preprint arXiv:2603.03596},
  year={2026}
}

@article{shi2025memoryvla,
  title={Memoryvla: Perceptual-cognitive memory in vision-language-action models for robotic manipulation},
  author={Shi, Hao and Xie, Bin and Liu, Yingfei and Sun, Lin and Liu, Fengrong and Wang, Tiancai and Zhou, Erjin and Fan, Haoqiang and Zhang, Xiangyu and Huang, Gao},
  journal={arXiv preprint arXiv:2508.19236},
  year={2025}
}

@article{li2026remem,
  title={ReMem-VLA: Empowering Vision-Language-Action Model with Memory via Dual-Level Recurrent Queries},
  author={Li, Hang and Shen, Fengyi and Chen, Dong and Yang, Liudi and Wang, Xudong and Shi, Jinkui and Bing, Zhenshan and Liu, Ziyuan and Knoll, Alois},
  journal={arXiv preprint arXiv:2603.12942},
  year={2026}
}

@inproceedings{rt22023arxiv,
  title={Rt-2: Vision-language-action models transfer web knowledge to robotic control},
  author={Zitkovich, Brianna and Yu, Tianhe and Xu, Sichun and Xu, Peng and Xiao, Ted and Xia, Fei and Wu, Jialin and Wohlhart, Paul and Welker, Stefan and Wahid, Ayzaan and others},
  booktitle={Conference on Robot Learning},
  pages={2165--2183},
  year={2023},
  organization={PMLR}
}

@article{liu2024rdt,
  title={Rdt-1b: a diffusion foundation model for bimanual manipulation},
  author={Liu, Songming and Wu, Lingxuan and Li, Bangguo and Tan, Hengkai and Chen, Huayu and Wang, Zhengyi and Xu, Ke and Su, Hang and Zhu, Jun},
  journal={arXiv preprint arXiv:2410.07864},
  year={2024}
}

@article{black2024pi0,
  title   = {{$\pi_0$}: A Vision-Language-Action Flow Model for General Robot Control},
  author  = {K. Black and N. Brown and D. Driess and A. Esmail and M. Equi and C. Finn and N. Fusai and L. Groom and K. Hausman and B. Ichter and others},
  journal = {arXiv preprint arXiv:2410.24164},
  year    = {2024}
}

@article{kim2024openvla,
  title={Openvla: An open-source vision-language-action model},
  author={Kim, Moo Jin and Pertsch, Karl and Karamcheti, Siddharth and Xiao, Ted and Balakrishna, Ashwin and Nair, Suraj and Rafailov, Rafael and Foster, Ethan and Lam, Grace and Sanketi, Pannag and others},
  journal={arXiv preprint arXiv:2406.09246},
  year={2024}
}

@article{kawaharazuka2025vision,
  title={Vision-language-action models for robotics: A review towards real-world applications},
  author={Kawaharazuka, Kento and Oh, Jihoon and Yamada, Jun and Posner, Ingmar and Zhu, Yuke},
  journal={IEEE Access},
  year={2025},
  publisher={IEEE}
}

@article{ma2024survey,
  title={A survey on vision-language-action models for embodied ai},
  author={Ma, Yueen and Song, Zixing and Zhuang, Yuzheng and Hao, Jianye and King, Irwin},
  journal={arXiv preprint arXiv:2405.14093},
  year={2024}
}

@article{sridhar2025memer,
  title={Memer: Scaling up memory for robot control via experience retrieval},
  author={Sridhar, Ajay and Pan, Jennifer and Sharma, Satvik and Finn, Chelsea},
  journal={arXiv preprint arXiv:2510.20328},
  year={2025}
}

@article{li2026roboclaw,
  title={RoboClaw: An Agentic Framework for Scalable Long-Horizon Robotic Tasks},
  author={Li, Ruiying and Zhou, Yunlang and Zhu, YuYao and Chen, Kylin and Wang, Jingyuan and Wang, Sukai and Hu, Kongtao and Yu, Minhui and Jiang, Bowen and Su, Zhan and others},
  journal={arXiv preprint arXiv:2603.11558},
  year={2026}
}

@article{zhao2023learning,
  title={Learning fine-grained bimanual manipulation with low-cost hardware},
  author={Zhao, Tony Z and Kumar, Vikash and Levine, Sergey and Finn, Chelsea},
  journal={arXiv preprint arXiv:2304.13705},
  year={2023}
}

@article{chi2025diffusion,
  title={Diffusion policy: Visuomotor policy learning via action diffusion},
  author={Chi, Cheng and Xu, Zhenjia and Feng, Siyuan and Cousineau, Eric and Du, Yilun and Burchfiel, Benjamin and Tedrake, Russ and Song, Shuran},
  journal={The International Journal of Robotics Research},
  volume={44},
  number={10-11},
  pages={1684--1704},
  year={2025},
  publisher={Sage Publications Sage UK: London, England}
}

@article{zheng2025x,
  title={X-vla: Soft-prompted transformer as scalable cross-embodiment vision-language-action model},
  author={Zheng, Jinliang and Li, Jianxiong and Wang, Zhihao and Liu, Dongxiu and Kang, Xirui and Feng, Yuchun and Zheng, Yinan and Zou, Jiayin and Chen, Yilun and Zeng, Jia and others},
  journal={arXiv preprint arXiv:2510.10274},
  year={2025}
}

@article{chen2026rmbench,
  title={RMBench: Memory-Dependent Robotic Manipulation Benchmark with Insights into Policy Design},
  author={Chen, Tianxing and Wang, Yuran and Li, Mingleyang and Qin, Yan and Shi, Hao and Li, Zixuan and Hu, Yifan and Zhang, Yingsheng and Wang, Kaixuan and Chen, Yue and others},
  journal={arXiv preprint arXiv:2603.01229},
  year={2026}
}

@article{zhang2026harness,
  title={Harness VLA: Steering Frozen VLAs into Reliable Manipulation Primitives via Memory-Guided Agents},
  author={Zhang, Yixian and Zhang, Huanming and Gao, Feng and Li, Xiao and Liu, Zhihao and Zhu, Chunyang and Qiu, Jiaxing and Yan, Yuchen and Liu, Jiyuan and Tang, Wenhao and others},
  journal={arXiv preprint arXiv:2607.08448},
  year={2026}
}

@article{ye2026world,
  title={World action models are zero-shot policies},
  author={Ye, Seonghyeon and Ge, Yunhao and Zheng, Kaiyuan and Gao, Shenyuan and Yu, Sihyun and Kurian, George and Indupuru, Suneel and Tan, You Liang and Zhu, Chuning and Xiang, Jiannan and others},
  journal={arXiv preprint arXiv:2602.15922},
  year={2026}
}

@article{kim2026cosmos,
  title={Cosmos policy: Fine-tuning video models for visuomotor control and planning},
  author={Kim, Moo Jin and Gao, Yihuai and Lin, Tsung-Yi and Lin, Yen-Chen and Ge, Yunhao and Lam, Grace and Liang, Percy and Song, Shuran and Liu, Ming-Yu and Finn, Chelsea and others},
  journal={arXiv preprint arXiv:2601.16163},
  year={2026}
}

@article{team2025gemini,
  title={Gemini robotics 1.5: Pushing the frontier of generalist robots with advanced embodied reasoning, thinking, and motion transfer},
  author={Team, Gemini Robotics and Abdolmaleki, Abbas and Abeyruwan, Saminda and Ainslie, Joshua and Alayrac, Jean-Baptiste and Arenas, Montserrat Gonzalez and Balakrishna, Ashwin and Batchelor, Nathan and Bewley, Alex and Bingham, Jeff and others},
  journal={arXiv preprint arXiv:2510.03342},
  year={2025}
}

@article{chen2026show,
  title={Show-harness: Just a vlm agent can play robots},
  author={Chen, Yanzhe and Bai, Zechen and Cao, Zhijun and Zeng, Wenzheng and Lin, Kevin Qinghong and Lin, Yiqi and Liang, Guoqiang and Ma, Kevin Yuchen and Huang, Qiming and Shou, Mike Zheng},
  journal={arXiv preprint arXiv:2609.10522},
  year={2026}
}

@article{liu2026hsc,
  title={HSC-VLA: Hierarchical Scene-Clearing for Robust Bimanual Manipulation in Dense Clutter},
  author={Liu, Zhen and Ning, Xinyu and Hu, Zhe and Xie, XinXin and Liu, Yitong and Pu, Zhongzhu},
  journal={arXiv preprint arXiv:2603.07484},
  year={2026}
}

@article{ding2026zetta,
  title={Zetta $\zeta$: An Efficient Closed-Loop Embodied Harness for Self-Evolving Physical Intelligence},
  author={Ding, Xin and Mi, Liang and Huang, Mingzhe and Wang, Zixuan and Zhang, Chao and Hao, Zixu and Chen, Fu and Li, Xiangyu and Zheng, Yikai and Guo, Yaoyu and others},
  journal={arXiv preprint arXiv:2608.16590},
  year={2026}
}

@article{fan2025long,
  title={Long-vla: Unleashing long-horizon capability of vision language action model for robot manipulation},
  author={Fan, Yiguo and Ding, Pengxiang and Bai, Shuanghao and Tong, Xinyang and Zhu, Yuyang and Lu, Hongchao and Dai, Fengqi and Zhao, Wei and Liu, Yang and Huang, Siteng and others},
  journal={arXiv preprint arXiv:2508.19958},
  year={2025}
}

@article{liu2026long,
  title={Long-horizon manipulation via trace-conditioned vla planning},
  author={Liu, Isabella and Cheng, An-Chieh and Yan, Rui and Chen, Geng and Qiu, Ri-Zhao and Zou, Xueyan and Yi, Sha and Yin, Hongxu and Wang, Xiaolong and Liu, Sifei},
  journal={arXiv preprint arXiv:2604.21924},
  year={2026}
}

@article{yan2026progressvla,
  title={Progressvla: Progress-guided diffusion policy for vision-language robotic manipulation},
  author={Yan, Hongyu and Li, Qiwei and Yang, Jiaolong and Mu, Yadong},
  journal={arXiv preprint arXiv:2603.27670},
  year={2026}
}

@inproceedings{xia2025phoenix,
  title={Phoenix: A motion-based self-reflection framework for fine-grained robotic action correction},
  author={Xia, Wenke and Feng, Ruoxuan and Wang, Dong and Hu, Di},
  booktitle={2025 IEEE/CVF Conference on Computer Vision and Pattern Recognition (CVPR)},
  pages={6981--6990},
  year={2025},
  organization={IEEE}
}

@article{yang2026physreflect,
  title={PhysReflect-VLA: Physical Feasibility and Self-Reflective Regulation for Reliable Vision-Language-Action Policies},
  author={Yang, Jiayu and Yang, Tao and Li, Weijun and Chang, Xiang and Chao, Fei and Shang, Changjing and Shen, Qiang},
  journal={arXiv preprint arXiv:2606.27146},
  year={2026}
}

@article{liu2026checkvla,
  title={CheckVLA: Execution-Time Verification with Action-Conditioned World Model for Long-Horizon Mobile Manipulation},
  author={Liu, Yushan and Sun, Peibo and Chao, Xintao and Yang, Zhenyang and Xie, Yifan and Zhang, Lingfeng and Li, Shoujie and Tang, Chenyu and Chen, Fang and Zhang, Xiao-Ping and others},
  journal={arXiv preprint arXiv:2607.26789},
  year={2026}
}

@article{dai2026robomme,
  title   = {RoboMME: Benchmarking and Understanding Memory for Robotic Generalist Policies},
  author  = {Dai, Yinpei and Fu, Hongze and Lee, Jayjun and Liu, Yuejiang
             and Zhang, Haoran and Yang, Jianing and Finn, Chelsea
             and Fazeli, Nima and Chai, Joyce},
  journal = {arXiv preprint arXiv:2603.04639},
  year    = {2026}
}

@article{fang2025sam2act,
  title={Sam2act: Integrating visual foundation model with a memory architecture for robotic manipulation},
  author={Fang, Haoquan and Grotz, Markus and Pumacay, Wilbert and Wang, Yi Ru and Fox, Dieter and Krishna, Ranjay and Duan, Jiafei},
  journal={arXiv preprint arXiv:2501.18564},
  year={2025}
}

@article{intelligence2025pi_,
  title={$\pi_{0.5}$: a Vision-Language-Action Model with Open-World Generalization},
  author={Intelligence, Physical and Black, Kevin and Brown, Noah and Darpinian, James and Dhabalia, Karan and Driess, Danny and Esmail, Adnan and Equi, Michael and Finn, Chelsea and Fusai, Niccolo and others},
  journal={arXiv preprint arXiv:2504.16054},
  year={2025}
}

@article{yang2026eventvla,
  title={EventVLA: Event-Driven Visual Evidence Memory for Long-Horizon Vision-Language-Action Policies},
  author={Yang, Ganlin and Tu, Zhangzheng and Yang, Yuqiang and Mao, Sitong and Dong, Junyi and Chen, Tianxing and Peng, Jiaqi and Xiong, Jing and Cao, Jiafei and Dai, Jifeng and others},
  journal={arXiv preprint arXiv:2606.20092},
  year={2026}
}

@article{zhou2023generalizable,
  title={Generalizable long-horizon manipulations with large language models},
  author={Zhou, Haoyu and Ding, Mingyu and Peng, Weikun and Tomizuka, Masayoshi and Shao, Lin and Gan, Chuang},
  journal={arXiv preprint arXiv:2310.02264},
  year={2023}
}

@inproceedings{li2023behavior,
  title={Behavior-1k: A benchmark for embodied ai with 1,000 everyday activities and realistic simulation},
  author={Li, Chengshu and Zhang, Ruohan and Wong, Josiah and Gokmen, Cem and Srivastava, Sanjana and Mart{\'\i}n-Mart{\'\i}n, Roberto and Wang, Chen and Levine, Gabrael and Lingelbach, Michael and Sun, Jiankai and others},
  booktitle={Conference on Robot Learning},
  pages={80--93},
  year={2023},
  organization={PMLR}
}

@article{han2026robocerebra,
  title={Robocerebra: A large-scale benchmark for long-horizon robotic manipulation evaluation},
  author={Han, Songhao and Qiu, Boxiang and Liao, Yue and Huang, Siyuan and Gao, Chen and Yan, Shuicheng and Liu, Si},
  journal={Advances in Neural Information Processing Systems},
  volume={38},
  year={2026}
}
\bibliographystyle{bibstyle}
\clearpage

\beginappendix

\section{Training Details}
\label{sec:app_training_details}

\subsection{RoboMME Training Details}
\label{sec:appendix_robomme_training}

For RoboMME, we follow the low-level policy training protocol of
RoboMME~\citep{dai2026robomme}. We initialize the low-level policy from
the $\pi_{0.5}$ backbone with the released LIBERO configuration and
train a single policy jointly over all 16 RoboMME tasks. Unless stated
otherwise, the optimization and policy configurations follow the
original RoboMME implementation. The complete training configuration is
summarized in Table~\ref{tab:robomme_training}. 
For RoboMME, we implement our method and reproduce all baselines using the officially released checkpoints. 

\begin{table}[h]
    \centering
    \caption{
    Training configuration of the low-level policy on RoboMME.
    We follow the original RoboMME training recipe and incorporate
    task-conditioned masked demonstrations into the joint multi-task
    training data.
    }
    \label{tab:robomme_training}

    \vspace{0.5ex}
    \small
    \setlength{\tabcolsep}{6pt}
    \renewcommand{\arraystretch}{1.05}

    \begin{tabular}{@{}ll@{}}
        \toprule
        Hyperparameter & Value \\
        \midrule

        Base policy
        & $\pi_{0.5}$ \\

        Policy configuration
        & LIBERO \\

        Number of training tasks
        & 16 \\

        Demonstrations per task
        & 100 \\

        Masked demonstrations per task
        & 20 \\

        Unmasked demonstrations per task
        & 80 \\

        Training scheme
        & Joint multi-task training \\

        \addlinespace[0.3ex]

        Action horizon
        & 20 \\


        Batch size
        & 64 \\

        Optimizer
        & AdamW \\

        $\beta_1$
        & 0.9 \\

        $\beta_2$
        & 0.95 \\

        Weight decay
        & 0 \\

        Gradient clip norm
        & 1.0 \\

        Learning-rate schedule
        & Constant \\

        Warmup ratio
        & 0.05 \\

        Learning rate
        & $5\times10^{-5}$ \\

        EMA decay
        & 0.999 \\

        Training steps
        & 80,000 \\

        Checkpoint interval
        & 10,000 \\

        Frozen module
        & SigLIP2 vision encoder \\

        Training hardware
        & $8\times$ NVIDIA A800 GPUs \\

        \bottomrule
    \end{tabular}
\end{table}

\paragraph{Masked training data.}
To expose the low-level policy to the task-conditioned visual filtering
used by Goal2Skill, we apply the masking procedure to 20 of the 100
training demonstrations for each task. The remaining 80 demonstrations
retain their original visual observations. Thus, the total number of
training demonstrations remains unchanged at 100 per task; the masked
and unmasked demonstrations are jointly used to train a single
multi-task policy over all 16 tasks. This procedure does not introduce any additional parameters into the base model.

\subsection{RMBench Training Details}
\label{sec:appendix_rmbench_training}

For RMBench, we follow the official $\pi_{0.5}$ fine-tuning
configuration used by the benchmark. The low-level policy is initialized
from the $\pi_{0.5}$ base checkpoint and trained using the benchmark
training recipe. For each evaluated task, 50 expert demonstrations are
used for policy training. The complete configuration is summarized in
Table~\ref{tab:rmbench_training}. For RMBench, we report the baseline results from~\cite{yang2026eventvla} and follow the same inference settings for a consistent comparison.

\begin{table}[t]
    \centering
    \caption{
    Training configuration of the low-level policy on RMBench.
    We follow the official $\pi_{0.5}$ training configuration and
    incorporate task-conditioned masked demonstrations into the
    training data.
    }
    \label{tab:rmbench_training}

    \vspace{0.5ex}
    \small
    \setlength{\tabcolsep}{6pt}
    \renewcommand{\arraystretch}{1.05}

    \begin{tabular}{@{}ll@{}}
        \toprule
        Hyperparameter & Value \\
        \midrule

        Base policy
        & $\pi_{0.5}$ \\

        Initialization
        & $\pi_{0.5}$ base checkpoint \\

        Demonstrations per task
        & 50 \\

        Masked demonstrations per task
        & 10 (20\%) \\

        Unmasked demonstrations per task
        & 40 (80\%) \\

        \addlinespace[0.3ex]

        Batch size
        & 64 \\

        Optimizer
        & AdamW \\

        $\beta_1$
        & 0.9 \\

        $\beta_2$
        & 0.95 \\

        Weight decay
        & $1\times10^{-10}$ \\

        Gradient clip norm
        & 1.0 \\

        Learning-rate schedule
        & Cosine decay \\

        Warmup steps
        & 1,000 \\

        Peak learning rate
        & $2.5\times10^{-5}$ \\

        Final learning rate
        & $2.5\times10^{-6}$ \\

        EMA decay
        & 0.99 \\

        Training steps
        & 20,000 \\

        Precision
        & bfloat16 \\

        \bottomrule
    \end{tabular}
\end{table}

\paragraph{Masked training data.}
For each RMBench task, we apply task-conditioned visual masking to
20\% of the training demonstrations. Specifically, 10 of the 50
demonstrations are processed with the masking procedure, while the
remaining 40 demonstrations retain their original visual observations.
The total number of demonstrations therefore remains unchanged, and
the masked and unmasked demonstrations are jointly used for policy
fine-tuning. This procedure does not introduce any additional parameters into the base model.

\subsection{High-Level Inference and Execution Details}
\label{sec:appendix_high_level}

\paragraph{VLM backend.}
All high-level reasoning modules use a single online VLM,
\texttt{qwen3.8-flash}, accessed through the OpenRouter API.
The model is used for task planning and replanning, semantic progress
verification, and reflective correction. During execution, the verifier
combines recent execution evidence with task memory and the expected
outcome of the active sub-task to determine its progress status.

\paragraph{Verification cadence.}
Following the verification procedure described in
Sec.~\ref{sec:progress_verification}, the active sub-task is verified
after every $n$ executed action chunks. A verification result is
classified as \textsc{completed}, \textsc{ongoing}, or \textsc{failed}.
A \textsc{completed} result advances execution to the next sub-task,
whereas \textsc{ongoing} retains the current sub-task and
\textsc{failed} triggers reflective correction.

\paragraph{Execution budgets.}
Each episode is subject to a global execution budget of
$T=1300$ environment steps. Each sub-task is assigned a per-attempt
execution budget of $\delta_k=320$ steps. If the active sub-task
exhausts this budget without verified completion, low-level execution
is interrupted and control returns to the high-level system for
reflection and recovery.

\paragraph{Recovery limits.}
Reflection selects among three recovery scopes:
\textsc{Retry}, \textsc{Adjust}, and \textsc{Replan}.
A sub-task may be attempted at most $K_{\mathrm{sub}}=3$ times.
After the planned sub-task sequence is exhausted, the verifier performs
a final check of the global goal. If the goal remains unverified,
reflection and replanning are allowed for at most three additional
rounds, subject to the remaining global execution budget.

\section{Case Study and Trajectory Analysis}
\subsection{Case Study on RoboMME Trajectories}
\label{sec:app_qualitative}

We further conduct qualitative analysis on six representative cases from
RoboMME to examine how verification and reflection contribute to
long-horizon task execution. The selected cases span six different tasks
and multiple difficulty levels, covering both
\emph{verification-driven progression} and
\emph{reflection-driven recovery}. In all six cases, the full Goal2Skill
framework successfully completes the task, whereas the compared
baselines fail because of incorrect intermediate execution, premature
task progression, or an inability to recover from an incomplete plan.
Together, these examples illustrate how verification grounds task
transitions in execution evidence, while reflection revises the remaining
plan when the recorded progress is inconsistent with the task outcome.

\begin{figure}[t]
    \centering
    \includegraphics[width=0.98\linewidth]
    {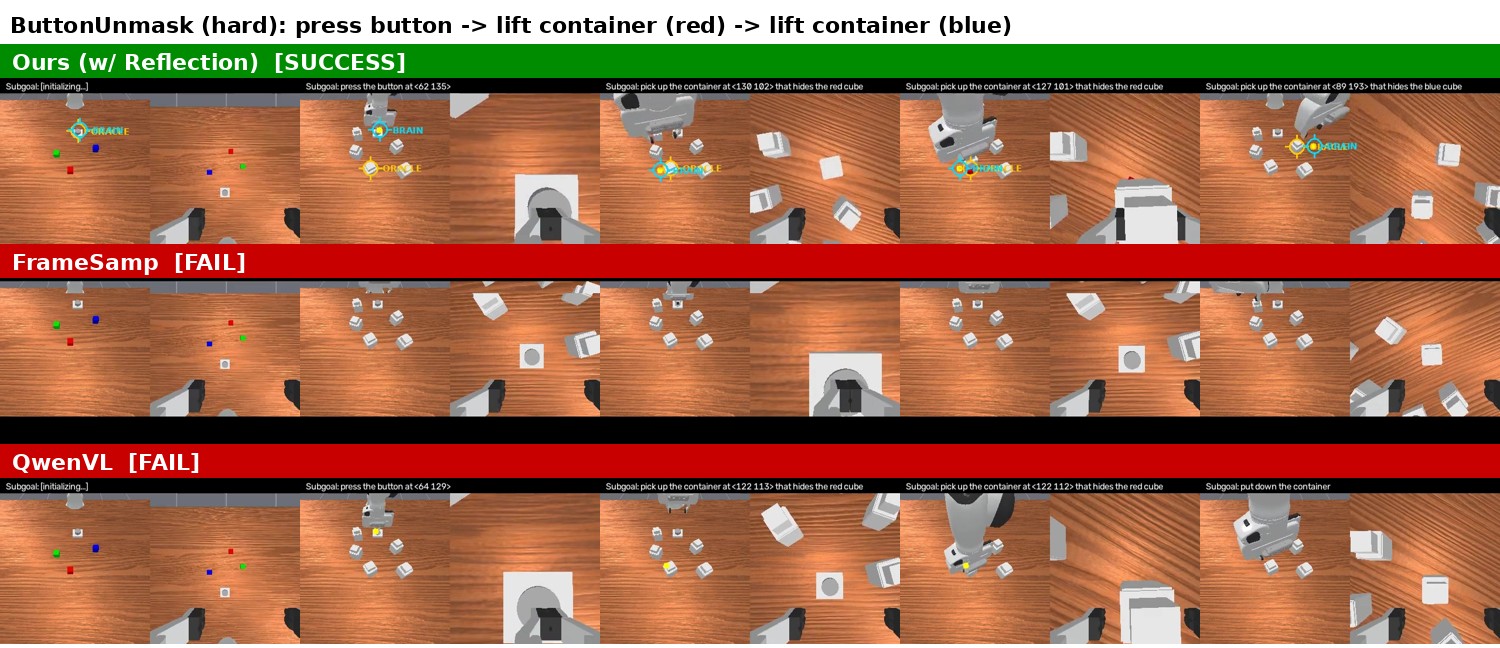}
    \caption{Reflection-driven recovery on \textsc{ButtonUnmask} (hard)
    from RoboMME. The initial plan terminates while the container covering
    the blue cube remains unopened. Reflection identifies the inconsistent
    completion record and revises the remaining execution.}
    \label{fig:case_buttonunmask}
\end{figure}

\paragraph{ButtonUnmask (hard).}
As shown in Figure~\ref{fig:case_buttonunmask}, the task requires the
robot to press a button and subsequently uncover the containers hiding
the red and blue cubes. Goal2Skill executes the initially planned stages,
but the episode remains unsuccessful after the plan is exhausted.
Reflection then revisits the recorded execution history and identifies a
false completion: the sub-task for uncovering the blue cube had been
marked as completed although its intended outcome was not confirmed.
The system therefore revises the unfinished execution and repeats the
corresponding uncovering operation, eventually completing the task.

In the same case, FrameSamp produces little effective task-directed
motion, while QwenVL generates intermediate sub-goals but fails to
physically complete the required manipulation, leaving the blue container
unopened. This example demonstrates how reflection can recover when a
recorded sub-task completion is inconsistent with the actual task state.

\begin{figure}[t]
    \centering
    \includegraphics[width=0.98\linewidth]
    {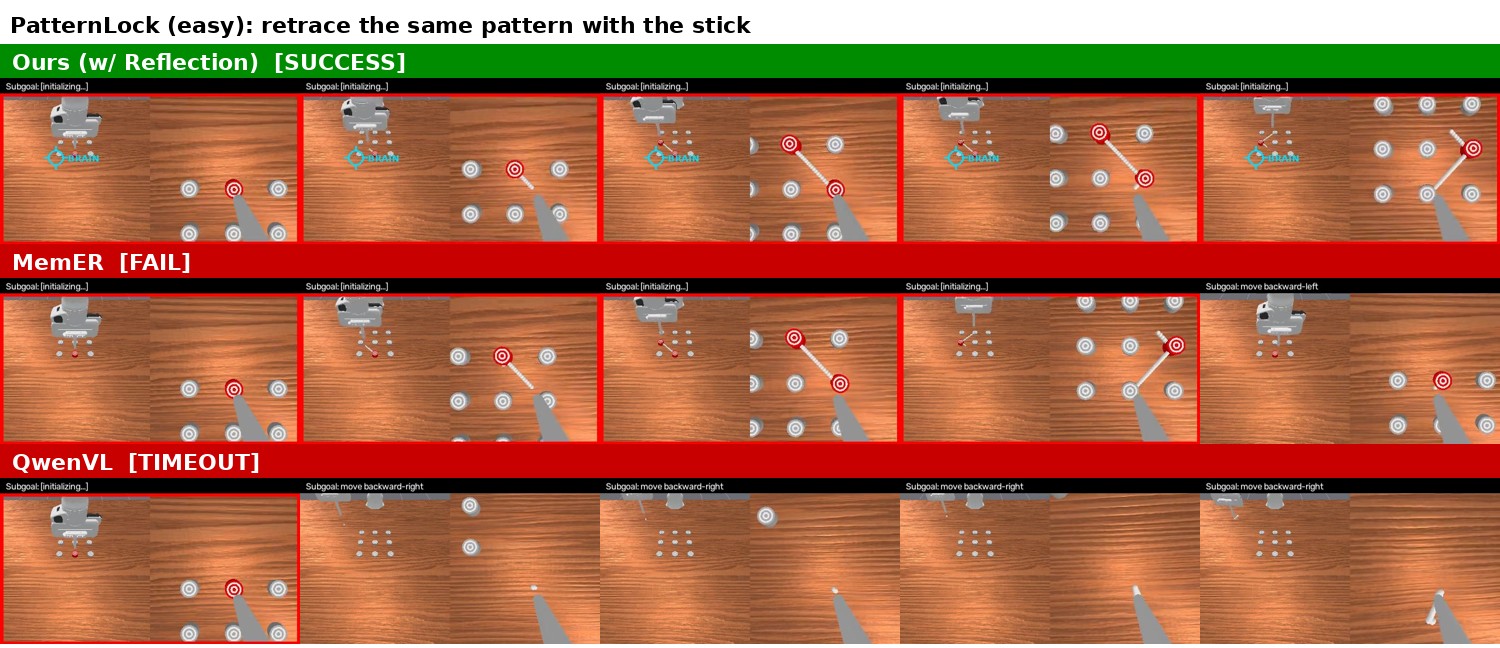}
    \caption{Reflection-driven correction on \textsc{PatternLock} (easy)
    from RoboMME. Reflection detects that the demonstrated trajectory
    remains incomplete after the initial plan terminates and replans the
    missing motion segment.}
    \label{fig:case_patternlock}
\end{figure}

\paragraph{PatternLock (easy).}
As illustrated in Figure~\ref{fig:case_patternlock}, the robot must use
the attached stick to reproduce a demonstrated trajectory. Although the
initial plan reaches its final stage, the global task remains incomplete.
Reflection examines the final recorded stage and determines that the
``move backward-left'' motion connects the red disk from the left-middle
region toward the bottom-middle region, but does not complete the target
pattern. The missing trajectory segment is subsequently replanned, and
Goal2Skill completes the required path.

By comparison, MemER deviates from the target trajectory after reaching
the central region and fails to complete the final segment, resulting in
failure. QwenVL makes limited effective progress and eventually times out
at the step budget. This case highlights the ability of reflection to
recover from execution that appears locally plausible but remains
globally incomplete.

\begin{figure}[t]
    \centering
    \includegraphics[width=0.98\linewidth]
    {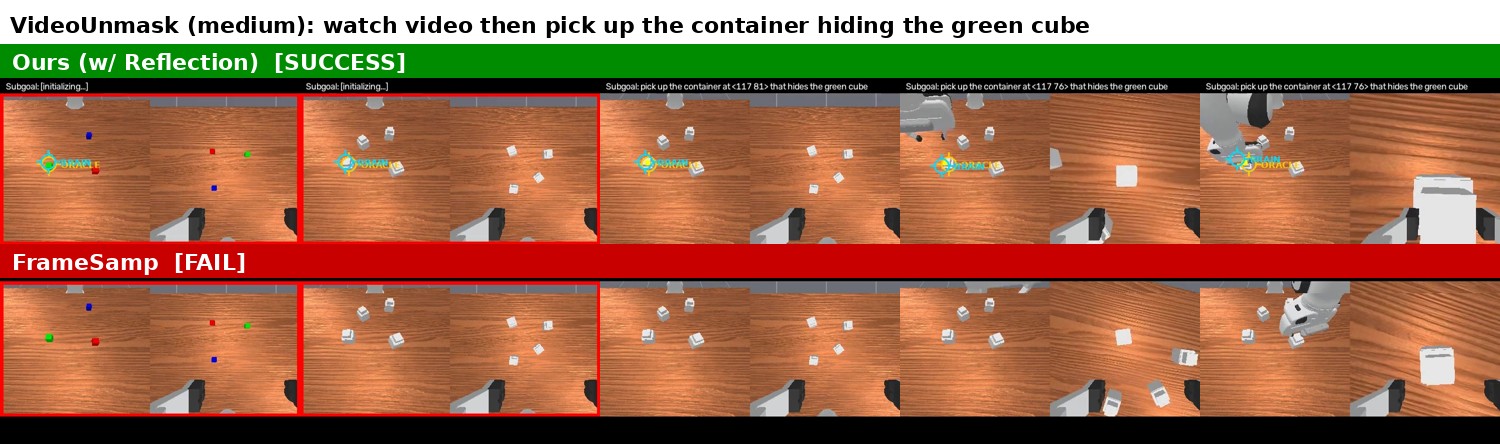}
    \caption{Reflection-driven recovery on \textsc{VideoUnmask} (medium)
    from RoboMME. A nominally completed container-lifting stage is
    invalidated by the unsuccessful terminal task state, triggering
    replanning until the target cube is uncovered.}
    \label{fig:case_videounmask}
\end{figure}

\paragraph{VideoUnmask (medium).}
As shown in Figure~\ref{fig:case_videounmask}, the task requires the
robot to infer the target from a demonstration video and uncover the
container hiding the green cube. After the container-lifting sub-task is
executed, the initial plan terminates without achieving the task goal.
Reflection identifies that the corresponding sub-task had been recorded
as successful despite the absence of task-level confirmation. The
completion record is therefore treated as invalid, and the unfinished
manipulation is replanned. After bounded replanning attempts, the execution
is corrected sufficiently to uncover the target cube.

FrameSamp, in contrast, fails to extract an effective target from the
demonstration and does not manipulate the correct container, resulting
in failure. This example illustrates how terminal task evidence can
expose a false intermediate completion, after which reflection can
redirect execution toward the unresolved operation.

\begin{figure}[t]
    \centering
    \includegraphics[width=0.98\linewidth]
    {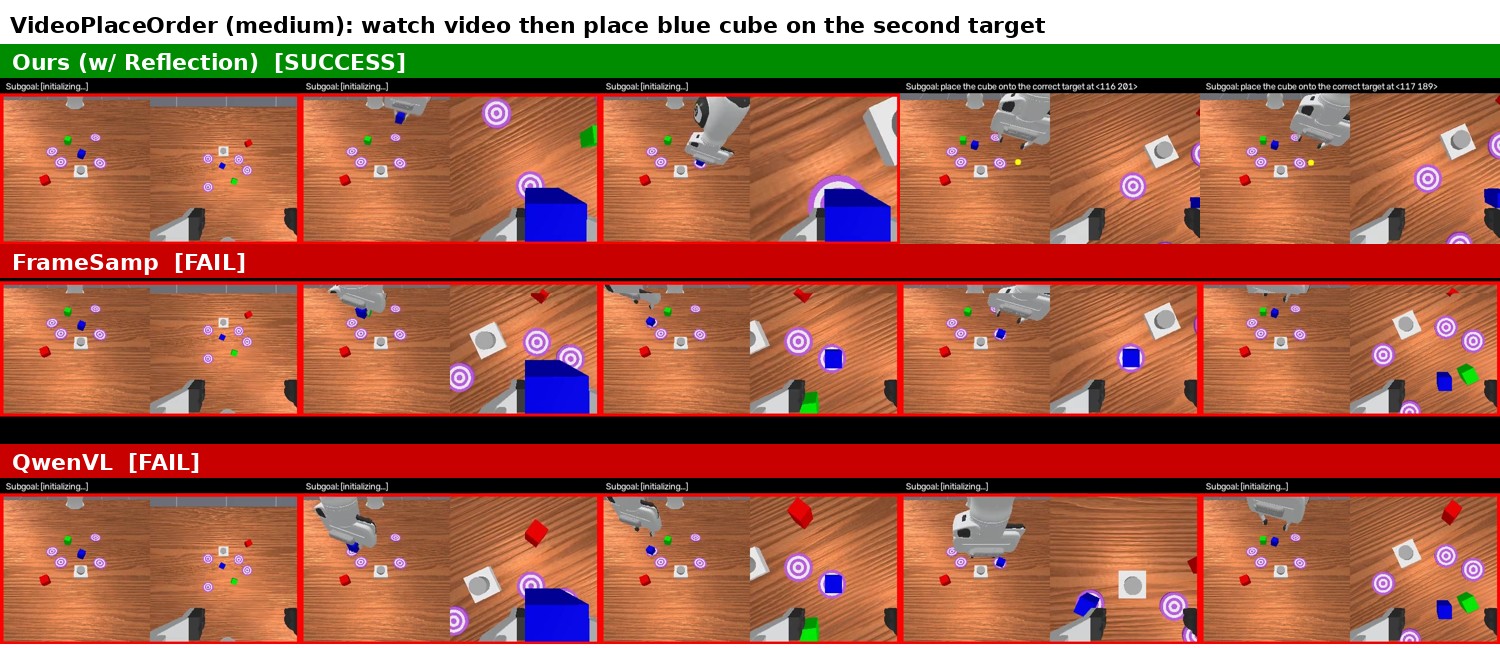}
    \caption{Verification-driven progression on
    \textsc{VideoPlaceOrder} (medium) from RoboMME. The verifier compares
    stage-level execution evidence and confirms that the blue cube reaches
    the required target location before advancing the plan.}
    \label{fig:case_videoplaceorder}
\end{figure}

\paragraph{VideoPlaceOrder (medium).}
As shown in Figure~\ref{fig:case_videoplaceorder}, the task requires the
robot to observe a demonstration video and place the blue cube at the
second target location used in the demonstration. During execution, the
verifier compares the evidence at the beginning and end of the placement
stage: the blue cube is initially located on the table and is later
observed at the intended target position. This evidence supports the
expected postcondition of the placement sub-task, allowing the verifier
to mark the stage as \textsc{completed} and safely advance the plan.

In contrast, FrameSamp and QwenVL fail to correctly resolve the
``second target position'' in this case, leading to an incorrect placement
or stalled execution. The example shows that explicit verification of
the physical outcome prevents task progression from relying solely on
whether a placement action has been issued.

\begin{figure}[t]
    \centering
    \includegraphics[width=0.98\linewidth]
    {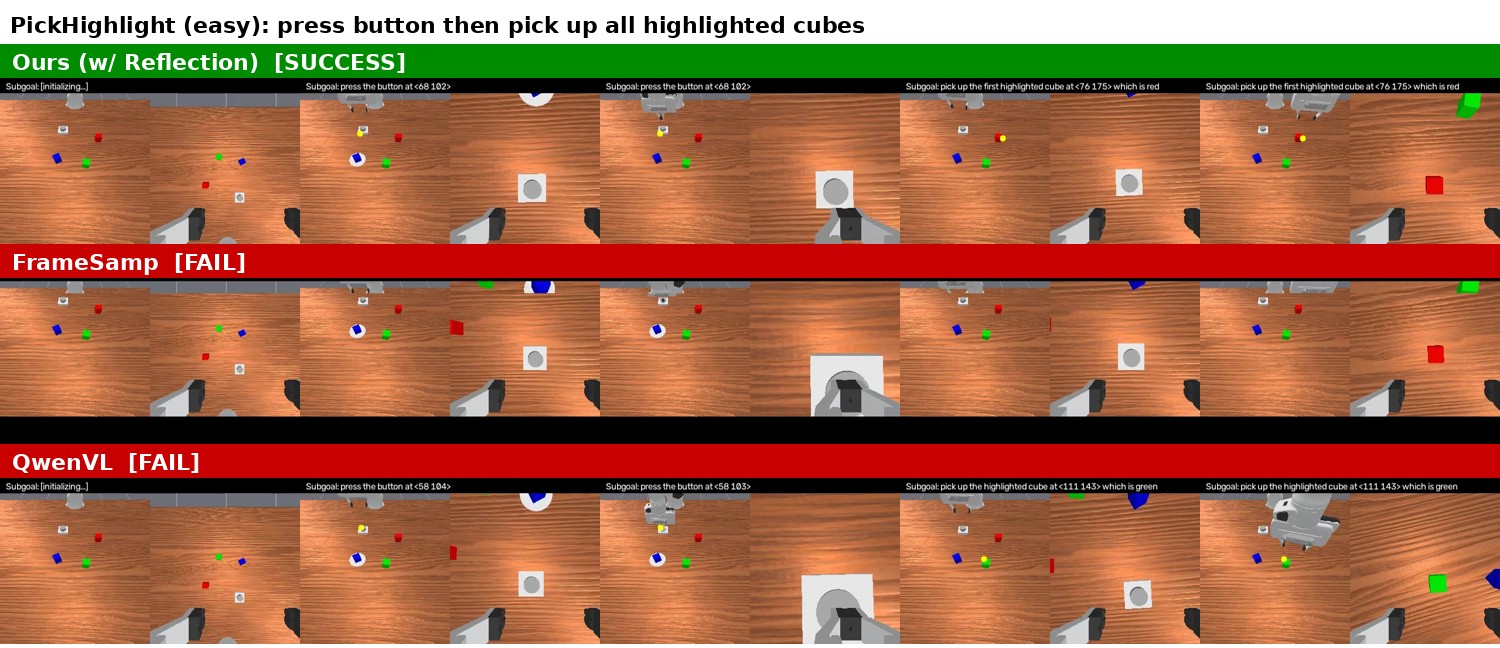}
    \caption{Verification- and reflection-guided plan expansion on
    \textsc{PickHighlight} (easy) from RoboMME. Verification detects the
    scene transition induced by the button press, after which reflection
    extends the plan to manipulate the newly highlighted cubes.}
    \label{fig:case_pickhighlight}
\end{figure}

\paragraph{PickHighlight (easy).}
As illustrated in Figure~\ref{fig:case_pickhighlight}, the task first
requires pressing a button and then picking all cubes marked by newly
appearing white highlights. Because the highlighted targets are not
available before the button interaction, the initial plan contains only
the button-press stage. Verification confirms that the button press has
produced the expected scene transition and that the highlight regions
have appeared. Reflection then triggers replanning
(\texttt{reason=pick\_highlight\_post\_button}) and introduces two
additional manipulation sub-tasks for the highlighted cubes, after which
the task is successfully completed.

FrameSamp and QwenVL, by comparison, terminate their effective execution
after the button interaction and do not proceed with the newly revealed
manipulation stage. This case demonstrates the complementary roles of
verification and reflection: verification establishes that the scene has
entered a new task state, while reflection uses the newly available
information to extend the unfinished plan.

\begin{figure}[t]
    \centering
    \includegraphics[width=0.98\linewidth]
    {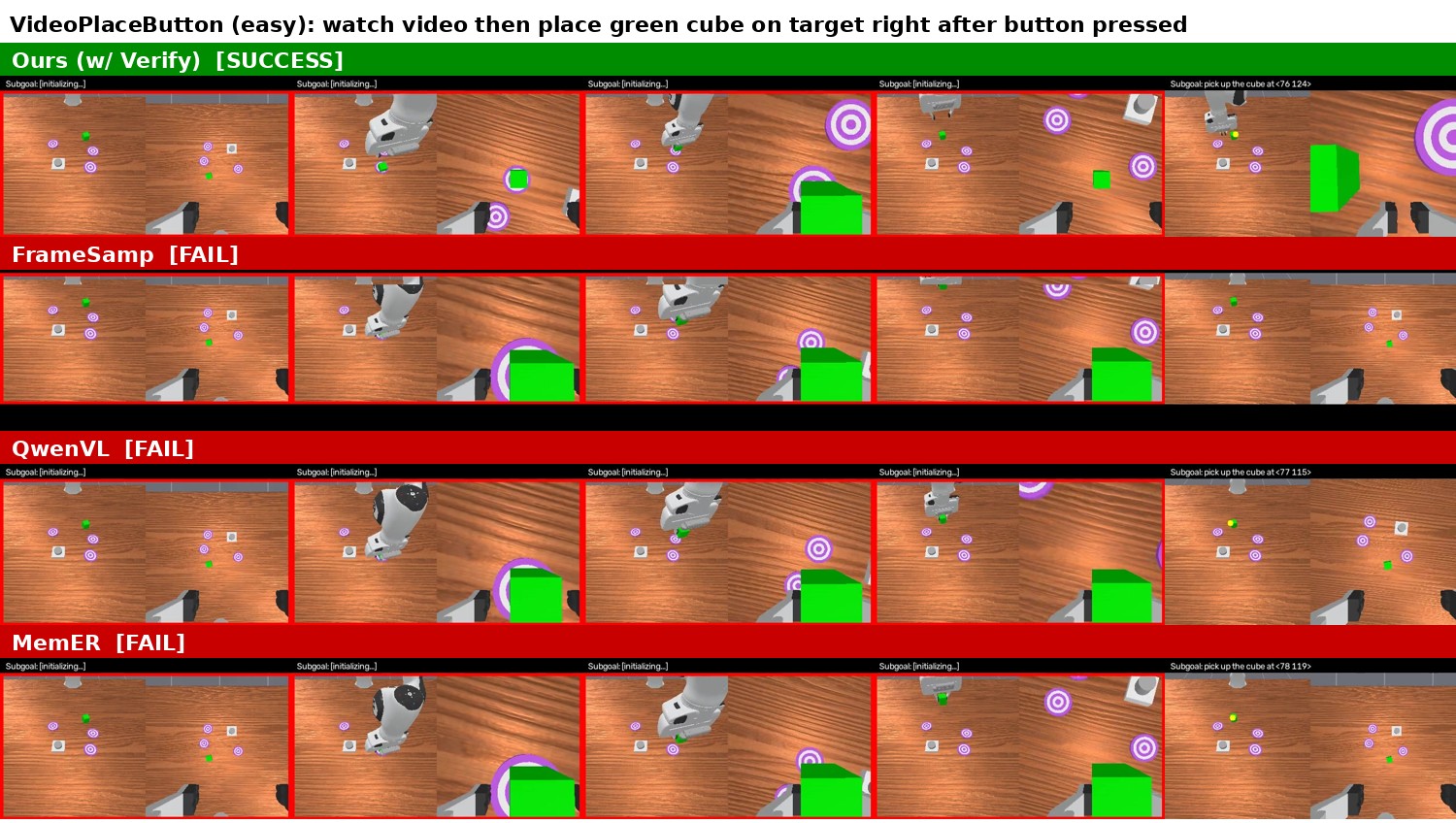}
    \caption{Verification-driven stage transition on
    \textsc{VideoPlaceButton} (easy) from RoboMME. The verifier confirms
    that the green cube has been securely grasped before execution
    proceeds to the placement stage.}
    \label{fig:case_videoplacebutton}
\end{figure}

\paragraph{VideoPlaceButton (easy).}
As shown in Figure~\ref{fig:case_videoplacebutton}, the task requires the
robot to interpret a demonstration video and place the green cube at the
target location after the button-triggered event. Before transitioning
from grasping to placement, the verifier compares the stage-start
observation with the current state. The green cube is initially located
on the table, whereas the current observation shows that it is securely
held by the gripper. This evidence satisfies the expected grasping
postcondition, allowing the plan to advance to the placement stage.

In the corresponding baseline executions, FrameSamp, QwenVL, and MemER
all complete the button interaction but fail during grasping. FrameSamp
positions the gripper away from the cube, QwenVL remains above the target
without establishing a grasp, and MemER keeps the gripper noticeably
separated from the object. These trajectories illustrate why explicit
verification of grasp completion is important before transitioning to
the subsequent placement stage.

\paragraph{Summary.}
The six RoboMME cases reveal two complementary benefits of the proposed
closed-loop framework. First, in \textsc{VideoPlaceOrder} and
\textsc{VideoPlaceButton}, verification grounds sub-task transitions in
observed execution outcomes. By requiring evidence that the expected
postcondition has actually been achieved, the system avoids advancing
the plan based only on nominal action execution.

Second, in \textsc{ButtonUnmask}, \textsc{PatternLock},
\textsc{VideoUnmask}, and \textsc{PickHighlight}, reflection addresses
mismatches between the current plan and the actual task state. It can
invalidate an incorrect completion record, recover missing operations,
or extend the unfinished plan when new task-relevant information becomes
available. Together, verification and reflection connect physical
execution outcomes with high-level task progression and targeted
recovery, providing a closed-loop mechanism for long-horizon
manipulation.

\subsection{Case Study on RMBench}

\begin{figure}[t]
    \centering
    \includegraphics[width=0.9\textwidth]{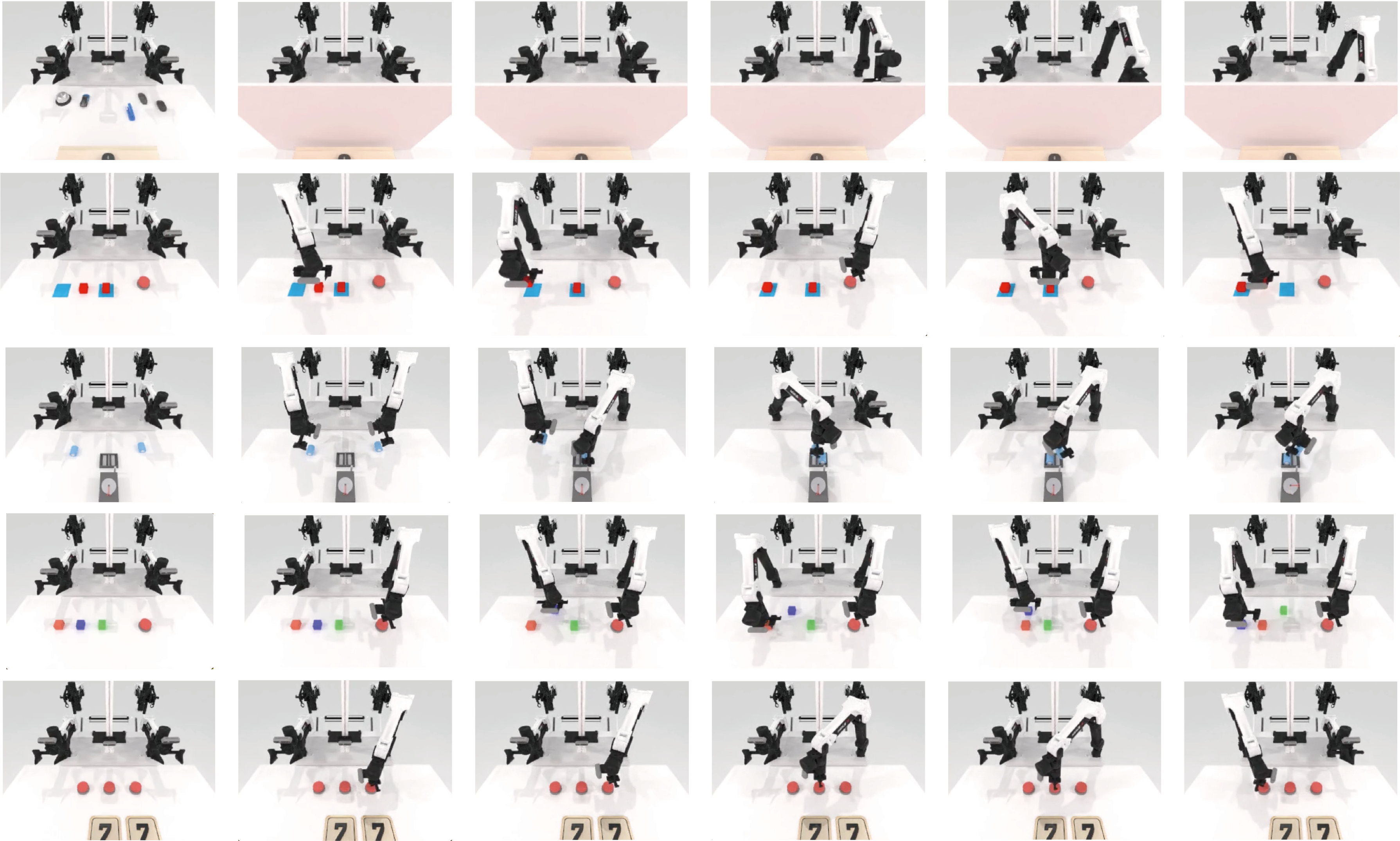}
    \caption{Sample trajectories of our framework on RMBench tasks. From top to bottom, the task sequences correspond to Observe and Pick Up, Rearrange Blocks, Battery Try, Blocks Ranking Try, and Press Button. Each row shows key frames from one rollout and illustrates that the method can complete representative memory-dependent manipulation tasks with different sequential interaction requirements.}
    \label{fig:qualitative_results}
\end{figure}

Figure~\ref{fig:qualitative_results} shows representative rollouts on the RMBench tasks. The examples show coherent progression across observation, rearrangement, trial-based interaction, ranking, and final goal verification. This visual evidence is consistent with the numerical results in Table~\ref{tab:rmbench_detailed}: the framework improves not only the final success statistic, but also illustrates coherent task progression across stages.

\section{Additional Experimental Results and Analysis}
\label{app:robomme_results}

\begin{table}[t]
    \centering
    \caption{
    Memory ablation on RMBench. We progressively remove components of structured memory to evaluate their contributions to tasks requiring target retention, intermediate state tracking, or multi-stage bookkeeping.
    }
    \label{tab:ablation_memory}

    \vspace{-2mm}
    \scriptsize
    \setlength{\tabcolsep}{2.8pt}
    \renewcommand{\arraystretch}{1.05}
    \begin{tabular}{l cccc}
        \toprule
        Setting &
        \shortstack[c]{Observe and Pick Up} &
        \shortstack[c]{Rearrange Blocks} &
        \shortstack[c]{Blocks Ranking Try} &
        Avg. \\
        \midrule
        Goal2Skill
        & 8\% & 38\% & 60\% & 35.3\% \\

        \quad $-\mathcal{E}_t$
        & 6\% & 36\% & 42\% & 28.0\% \\

        \quad $-\mathcal{E}_t-\mathcal{W}_t$
        & 8\% & 21\% & 54\% & 27.7\% \\

        \quad $-\mathcal{E}_t-\mathcal{W}_t-\mathcal{H}_t$
        & 4\% & 10\% & 6\% & 6.7\% \\
        \bottomrule
    \end{tabular}
\end{table}

\subsection{Ablation Study}
\label{sec:exp_ablation}

We conduct targeted ablations on RMBench tasks to examine the contributions of structured memory and reflective correction. The task subsets are chosen to emphasize the capabilities associated with each component, including long-horizon state tracking and recovery from failed execution. 

\noindent\textbf{Structured memory.}
As shown in Table~\ref{tab:ablation_memory}, the full Goal2Skill model achieves an average success rate of 35.3\%. Removing the error register $\mathcal{E}_t$ reduces the average to 28.0\%, with \emph{Blocks Ranking Try} decreasing from 60\% to 42\%. This suggests that retaining failure and recovery information contributes to robust long-horizon behavior, particularly when earlier execution outcomes are relevant to later decisions. Further removing working memory $\mathcal{W}_t$ results in an average success rate of 27.7\%. Although the overall average changes only slightly, the effect varies across tasks: \emph{Rearrange Blocks} decreases from 36\% to 21\%, while \emph{Blocks Ranking Try} increases from 42\% to 54\%. This indicates that working memory is particularly useful for tracking evolving intermediate states, but its contribution is task dependent. Finally, removing episodic history $\mathcal{H}_t$ as well reduces the model without any structured memory, causing the average success rate to drop sharply from 27.7\% to 6.7\%. 
This setting even falls slightly below the $\pi_{0.5}$ baseline, likely because the planner, without any record of completed sub-tasks, can no longer track progress and tends to repeat finished steps or advance prematurely, issuing sub-task instructions that conflict with the actual scene and disrupt the low-level policy. This further indicates that task-level planning is only beneficial when grounded in persistent memory.
The largest decrease occurs on \emph{Blocks Ranking Try}, from 54\% to 6\%, highlighting the importance of persistent chronological task history for delayed target recall and multi-stage bookkeeping. Overall, the results show that the three memory components provide complementary functionality, with the complete structured memory achieving the strongest overall performance.

\noindent\textbf{Reflection.}
Table~\ref{tab:ablation_recovery} evaluates the contribution of reflective correction. The full model achieves 28.0\% on these two tasks, reaching 46\% on \emph{Battery Try} and 10\% on \emph{Press Button}. Removing reflection reduces the average to 17.5\%, with \emph{Battery Try} decreasing from 46\% to 30\% and \emph{Press Button} from 10\% to 5\%. Although failures are still detected, they can only be addressed by re-executing the same sub-task, without diagnosing the cause or revising the plan. This suggests that verification signals are most effective when translated into explicit corrective actions such as adjusting execution or replanning.


\begin{table}[t]
    \centering
    \caption{
    Reflection ablation on RMBench.
    }
    \label{tab:ablation_recovery}

    \vspace{-2mm}
    \scriptsize
    \setlength{\tabcolsep}{3.2pt}
    \renewcommand{\arraystretch}{1.05}
    \begin{tabular}{l ccc}
        \toprule
        Setting &
        \shortstack[c]{Battery Try} &
        \shortstack[c]{Press Button} &
        Avg. \\
        \midrule
        Goal2Skill
        & 46\% & 10\% & 28.0\% \\

        \quad $-\mathrm{Reflection}$
        & 30\% & 5\% & 17.5\% \\

        \bottomrule
    \end{tabular}
\end{table}

\subsection{Complete RoboMME Results}
\label{app:robomme_complete}

\begin{table}[t]
    \centering
    \caption{
    Complete per-task success rates (\%) on all 16 RoboMME
    tasks~\citep{dai2026robomme}.
    Best and second-best results among non-oracle methods are shown in
    \textbf{bold} and \underline{underline}, respectively.
    GroundSG+Oracle is included as an oracle reference.
    }
    \label{tab:robomme_full}

    \vspace{0.5ex}
    \scriptsize
    \setlength{\tabcolsep}{2.0pt}
    \renewcommand{\arraystretch}{1.08}

    \resizebox{\textwidth}{!}{%
    \begin{tabular}{@{}l*{17}{c}@{}}
        \toprule

        \multirow{2}{*}{Method}
        & \multicolumn{4}{c}{Counting}
        & \multicolumn{4}{c}{Permanence}
        & \multicolumn{4}{c}{Reference}
        & \multicolumn{4}{c}{Imitation}
        & \multirow{2}{*}{Overall}
        \\

        \cmidrule(lr){2-5}
        \cmidrule(lr){6-9}
        \cmidrule(lr){10-13}
        \cmidrule(lr){14-17}

        &
        \shortstack[c]{Bin\\Fill}
        &
        \shortstack[c]{Pick\\Xtimes}
        &
        \shortstack[c]{Swing\\Xtimes}
        &
        \shortstack[c]{Stop\\Cube}
        &
        \shortstack[c]{Video\\Unmask}
        &
        \shortstack[c]{Video\\Unmask\\Swap}
        &
        \shortstack[c]{Button\\Unmask}
        &
        \shortstack[c]{Button\\Unmask\\Swap}
        &
        \shortstack[c]{Pick\\Highlight}
        &
        \shortstack[c]{Video\\Repick}
        &
        \shortstack[c]{Video\\Place\\Button}
        &
        \shortstack[c]{Video\\Place\\Order}
        &
        \shortstack[c]{Move\\Cube}
        &
        \shortstack[c]{Insert\\Peg}
        &
        \shortstack[c]{Pattern\\Lock}
        &
        \shortstack[c]{Route\\Stick}
        &
        \\
        \midrule

        \rowcolor{gray!15}
        \multicolumn{18}{@{}l}{\textit{Base Policies}} \\

        $\pi_{0.5}$
        & 30.0
        & 42.0
        & 36.0
        & \underline{6.0}
        & 20.0
        & 18.0
        & 22.0
        & 6.0
        & 12.0
        & 0.0
        & 32.0
        & 26.0
        & 26.0
        & \underline{2.0}
        & 2.0
        & 4.0
        & 17.8
        \\

        $\pi_{0.5}$ w/ past actions
        & 26.0
        & 58.0
        & 26.0
        & 4.0
        & 30.0
        & 20.0
        & 24.0
        & 16.0
        & 12.0
        & 8.0
        & 24.0
        & 18.0
        & 34.0
        & \underline{2.0}
        & 4.0
        & 6.0
        & 19.5
        \\

        \midrule

        \rowcolor{gray!15}
        \multicolumn{18}{@{}l}{\textit{Memory-Augmented VLA Policies}} \\

        MME-VLA (FrameSamp)
        & 38.0
        & 82.0
        & \textbf{94.0}
        & \textbf{50.0}
        & 38.0
        & \underline{28.0}
        & 26.0
        & 12.0
        & 24.0
        & 32.0
        & \textbf{54.0}
        & 28.0
        & 48.0
        & \underline{2.0}
        & \textbf{54.0}
        & \textbf{66.0}
        & 42.3
        \\

        MemER
        & \textbf{66.0}
        & 78.0
        & \underline{60.0}
        & 0.0
        & \underline{86.0}
        & \textbf{36.0}
        & \textbf{78.0}
        & 24.0
        & \textbf{82.0}
        & 32.0
        & 16.0
        & \underline{36.0}
        & \textbf{82.0}
        & \underline{2.0}
        & 14.0
        & 6.0
        & \underline{43.6}
        \\

        SAM2Act+
        & 40.0
        & 76.0
        & 26.0
        & 0.0
        & 28.0
        & 18.0
        & 32.0
        & \underline{26.0}
        & 18.0
        & 6.0
        & 24.0
        & 20.0
        & 30.0
        & 0.0
        & 0.0
        & 0.0
        & 21.5
        \\

        \midrule

        \rowcolor{gray!15}
        \multicolumn{18}{@{}l}{\textit{Dual Systems}} \\

        GroundSG+QwenVL
        & 48.0
        & \underline{90.0}
        & 6.0
        & 0.0
        & \textbf{88.0}
        & \underline{28.0}
        & 26.0
        & 10.0
        & 24.0
        & \underline{34.0}
        & 48.0
        & \underline{36.0}
        & \underline{68.0}
        & \textbf{6.0}
        & 8.0
        & 4.0
        & 32.8
        \\

        \textbf{Goal2Skill}
        & \underline{58.0}
        & \textbf{92.0}
        & 56.0
        & 2.0
        & 84.0
        & 24.0
        & \underline{70.0}
        & \textbf{50.0}
        & \underline{76.0}
        & \textbf{60.0}
        & \underline{52.0}
        & \textbf{90.0}
        & \underline{68.0}
        & \underline{2.0}
        & \underline{32.0}
        & \underline{8.0}
        & \textbf{51.5}
        \\

        \midrule

        \rowcolor{gray!15}
        \multicolumn{18}{@{}l}{\textit{Oracle Reference}} \\

        GroundSG+Oracle
        & 90.0
        & 100.0
        & 100.0
        & 44.0
        & 98.0
        & 96.0
        & 98.0
        & 86.0
        & 52.0
        & 98.0
        & 100.0
        & 98.0
        & 90.0
        & 10.0
        & 100.0
        & 64.0
        & 82.8
        \\

        \bottomrule
    \end{tabular}%
    }
\end{table}

\paragraph{Complete RoboMME results.}
Table~\ref{tab:robomme_full} shows that the trend in the main-text
comparison extends to the full RoboMME benchmark. Goal2Skill achieves
the highest overall success rate among non-oracle methods at 51.5\%,
followed by MemER at 43.6\% and MME-VLA (FrameSamp) at 42.3\%.
Goal2Skill also ranks among the top two non-oracle methods on 12 of the
16 tasks.

The results differ across memory categories. Goal2Skill averages 57.0\%
on \emph{Permanence} and 69.5\% on \emph{Reference}. The gain is most
pronounced on \emph{Reference}, where MemER reaches 41.5\%. These tasks
require information observed earlier in the episode to remain available
for later decisions. This pattern is consistent with the role of
persistent task memory in Goal2Skill.

On \emph{Counting}, Goal2Skill reaches 52.0\%, while FrameSamp achieves
66.0\%. On \emph{Imitation}, Goal2Skill reaches 27.5\%, compared with
42.5\% for FrameSamp. The task-level results show a similar division:
Goal2Skill is strong when success depends on maintaining task state over
time, while methods with strong perceptual history remain competitive on
tasks such as \emph{SwingXtimes}, \emph{PatternLock}, and
\emph{RouteStick}. Overall, the full benchmark shows that the main gain
of Goal2Skill comes from long-horizon state tracking and coordination.


\subsection{Difficulty-Stratified RoboMME Results}
\label{app:robomme_difficulty}


\begin{table}[t]
    \centering
    \caption{
    Difficulty-stratified evaluation on the Counting and Permanence
    tasks of RoboMME~\citep{dai2026robomme}.
    We report success rates (\%) for Easy, Medium, and Hard episodes.
    Each task contains 50 test episodes, including 26 Easy, 12 Medium,
    and 12 Hard episodes.
    GroundSG+Oracle is included as an oracle reference.
    }
    \label{tab:robomme_difficulty_cp}

    \vspace{0.5ex}
    \small
    \setlength{\tabcolsep}{3.0pt}
    \renewcommand{\arraystretch}{1.05}

    \resizebox{\textwidth}{!}{%
    \begin{tabular}{@{}llcccccccc@{}}
        \toprule

        \multirow{2}{*}{Method}
        & \multirow{2}{*}{Difficulty}
        & \multicolumn{4}{c}{Counting}
        & \multicolumn{4}{c}{Permanence}
        \\

        \cmidrule(lr){3-6}
        \cmidrule(lr){7-10}

        &
        &
        \shortstack[c]{Bin\\Fill}
        &
        \shortstack[c]{Pick\\Xtimes}
        &
        \shortstack[c]{Swing\\Xtimes}
        &
        \shortstack[c]{Stop\\Cube}
        &
        \shortstack[c]{Video\\Unmask}
        &
        \shortstack[c]{Video\\Unmask\\Swap}
        &
        \shortstack[c]{Button\\Unmask}
        &
        \shortstack[c]{Button\\Unmask\\Swap}
        \\

        \midrule

        \rowcolor{gray!15}
        \multicolumn{10}{@{}l}{\textit{Base Policies}} \\

        \multirow{3}{*}{$\pi_{0.5}$}
        & Easy
        & 38.5 & 46.2 & 34.6 & 7.7
        & 23.1 & 19.2 & 23.1 & 7.7
        \\
        & Medium
        & 25.0 & 41.7 & 41.7 & 8.3
        & 16.7 & 16.7 & 25.0 & 8.3
        \\
        & Hard
        & 16.7 & 33.3 & 33.3 & 0.0
        & 16.7 & 16.7 & 16.7 & 0.0
        \\

        \addlinespace[0.35ex]

        \multirow{3}{*}{
            \shortstack[l]{
                $\pi_{0.5}$ w/\\
                past actions
            }
        }
        & Easy
        & 34.6 & 61.5 & 26.9 & 3.8
        & 34.6 & 19.2 & 26.9 & 19.2
        \\
        & Medium
        & 25.0 & 66.7 & 25.0 & 8.3
        & 25.0 & 25.0 & 25.0 & 16.7
        \\
        & Hard
        & 8.3 & 41.7 & 25.0 & 0.0
        & 25.0 & 16.7 & 16.7 & 8.3
        \\

        \midrule

        \rowcolor{gray!15}
        \multicolumn{10}{@{}l}{\textit{Memory-Augmented VLA Policies}} \\

        \multirow{3}{*}{MME-VLA (FrameSamp)}
        & Easy
        & 53.8 & 92.3 & 96.2 & 53.8
        & 46.2 & 34.6 & 34.6 & 15.4
        \\
        & Medium
        & 41.7 & 100.0 & 100.0 & 41.7
        & 25.0 & 33.3 & 33.3 & 16.7
        \\
        & Hard
        & 0.0 & 41.7 & 83.3 & 50.0
        & 33.3 & 8.3 & 0.0 & 0.0
        \\

        \addlinespace[0.35ex]

        \multirow{3}{*}{MemER}
        & Easy
        & 84.6 & 96.2 & 53.8 & 0.0
        & 96.2 & 34.6 & 84.6 & 34.6
        \\
        & Medium
        & 58.3 & 83.3 & 83.3 & 0.0
        & 91.7 & 50.0 & 91.7 & 25.0
        \\
        & Hard
        & 33.3 & 33.3 & 50.0 & 0.0
        & 58.3 & 25.0 & 50.0 & 0.0
        \\

        \addlinespace[0.35ex]

        \multirow{3}{*}{SAM2Act+}
        & Easy
        & 50.0 & 80.8 & 26.9 & 0.0
        & 30.8 & 19.2 & 34.6 & 30.8
        \\
        & Medium
        & 33.3 & 83.3 & 25.0 & 0.0
        & 25.0 & 16.7 & 33.3 & 25.0
        \\
        & Hard
        & 25.0 & 58.3 & 25.0 & 0.0
        & 25.0 & 16.7 & 25.0 & 16.7
        \\

        \midrule

        \rowcolor{gray!15}
        \multicolumn{10}{@{}l}{\textit{Dual Systems}} \\

        \multirow{3}{*}{GroundSG+QwenVL}
        & Easy
        & 76.9 & 92.3 & 7.7 & 0.0
        & 100.0 & 26.9 & 38.5 & 11.5
        \\
        & Medium
        & 16.7 & 91.7 & 0.0 & 0.0
        & 91.7 & 33.3 & 25.0 & 16.7
        \\
        & Hard
        & 16.7 & 83.3 & 8.3 & 0.0
        & 58.3 & 25.0 & 0.0 & 0.0
        \\

        \addlinespace[0.35ex]

        \multirow{3}{*}{\textbf{Goal2Skill}}
        & Easy
        & 73.1 & 96.2 & 73.1 & 3.8
        & 92.3 & 34.6 & 84.6 & 65.4
        \\
        & Medium
        & 50.0 & 91.7 & 50.0 & 0.0
        & 83.3 & 16.7 & 66.7 & 41.7
        \\
        & Hard
        & 33.3 & 83.3 & 25.0 & 0.0
        & 66.7 & 8.3 & 41.7 & 25.0
        \\

        \midrule

        \rowcolor{gray!15}
        \multicolumn{10}{@{}l}{\textit{Oracle Reference}} \\

        \multirow{3}{*}{GroundSG+Oracle}
        & Easy
        & 96.2 & 100.0 & 100.0 & 50.0
        & 100.0 & 96.2 & 96.2 & 92.3
        \\
        & Medium
        & 91.7 & 100.0 & 100.0 & 41.7
        & 100.0 & 100.0 & 100.0 & 91.7
        \\
        & Hard
        & 75.0 & 100.0 & 100.0 & 33.3
        & 91.7 & 91.7 & 100.0 & 66.7
        \\

        \bottomrule
    \end{tabular}%
    }
\end{table}


\begin{table}[t]
    \centering
    \caption{
    Difficulty-stratified evaluation on the Reference and Imitation
    tasks of RoboMME~\citep{dai2026robomme}.
    We report success rates (\%) for Easy, Medium, and Hard episodes.
    Each task contains 50 test episodes, including 26 Easy, 12 Medium,
    and 12 Hard episodes.
    GroundSG+Oracle is included as an oracle reference.
    }
    \label{tab:robomme_difficulty_ri}

    \vspace{0.3ex}
    \small
    \setlength{\tabcolsep}{3.0pt}
    \renewcommand{\arraystretch}{0.95}

    \resizebox{\textwidth}{!}{%
    \begin{tabular}{@{}llcccccccc@{}}
        \toprule

        \multirow{2}{*}{Method}
        & \multirow{2}{*}{Difficulty}
        & \multicolumn{4}{c}{Reference}
        & \multicolumn{4}{c}{Imitation}
        \\

        \cmidrule(lr){3-6}
        \cmidrule(lr){7-10}

        &
        &
        \shortstack[c]{Pick\\Highlight}
        &
        \shortstack[c]{Video\\Repick}
        &
        \shortstack[c]{Video\\Place\\Button}
        &
        \shortstack[c]{Video\\Place\\Order}
        &
        \shortstack[c]{Move\\Cube}
        &
        \shortstack[c]{Insert\\Peg}
        &
        \shortstack[c]{Pattern\\Lock}
        &
        \shortstack[c]{Route\\Stick}
        \\

        \midrule

        \rowcolor{gray!15}
        \multicolumn{10}{@{}l}{\textit{Base Policies}} \\

        \multirow{3}{*}{$\pi_{0.5}$}
        & Easy
        & 15.4 & 0.0 & 34.6 & 26.9
        & 26.9 & 3.8 & 3.8 & 7.7
        \\
        & Medium
        & 8.3 & 0.0 & 33.3 & 25.0
        & 25.0 & 0.0 & 0.0 & 0.0
        \\
        & Hard
        & 8.3 & 0.0 & 25.0 & 25.0
        & 25.0 & 0.0 & 0.0 & 0.0
        \\

        \addlinespace[0.15ex]

        \multirow{3}{*}{
            \shortstack[l]{
                $\pi_{0.5}$ w/\\
                past actions
            }
        }
        & Easy
        & 15.4 & 7.7 & 26.9 & 19.2
        & 34.6 & 3.8 & 3.8 & 7.7
        \\
        & Medium
        & 8.3 & 8.3 & 25.0 & 16.7
        & 33.3 & 0.0 & 8.3 & 0.0
        \\
        & Hard
        & 8.3 & 8.3 & 16.7 & 16.7
        & 33.3 & 0.0 & 0.0 & 8.3
        \\

        \midrule

        \rowcolor{gray!15}
        \multicolumn{10}{@{}l}{\textit{Memory-Augmented VLA Policies}} \\

        \multirow{3}{*}{MME-VLA (FrameSamp)}
        & Easy
        & 23.1 & 46.2 & 65.4 & 34.6
        & 50.0 & 0.0 & 80.8 & 92.3
        \\
        & Medium
        & 33.3 & 25.0 & 58.3 & 25.0
        & 41.7 & 8.3 & 50.0 & 41.7
        \\
        & Hard
        & 16.7 & 8.3 & 25.0 & 16.7
        & 50.0 & 0.0 & 0.0 & 33.3
        \\

        \addlinespace[0.15ex]

        \multirow{3}{*}{MemER}
        & Easy
        & 96.2 & 26.9 & 23.1 & 34.6
        & 73.1 & 0.0 & 26.9 & 11.5
        \\
        & Medium
        & 83.3 & 33.3 & 16.7 & 41.7
        & 91.7 & 8.3 & 0.0 & 0.0
        \\
        & Hard
        & 50.0 & 41.7 & 0.0 & 33.3
        & 91.7 & 0.0 & 0.0 & 0.0
        \\

        \addlinespace[0.15ex]

        \multirow{3}{*}{SAM2Act+}
        & Easy
        & 23.1 & 7.7 & 26.9 & 23.1
        & 26.9 & 0.0 & 0.0 & 0.0
        \\
        & Medium
        & 16.7 & 8.3 & 25.0 & 16.7
        & 33.3 & 0.0 & 0.0 & 0.0
        \\
        & Hard
        & 8.3 & 0.0 & 16.7 & 16.7
        & 33.3 & 0.0 & 0.0 & 0.0
        \\

        \midrule

        \rowcolor{gray!15}
        \multicolumn{10}{@{}l}{\textit{Dual Systems}} \\

        \multirow{3}{*}{GroundSG+QwenVL}
        & Easy
        & 38.5 & 30.8 & 50.0 & 42.3
        & 57.7 & 11.5 & 15.4 & 7.7
        \\
        & Medium
        & 16.7 & 50.0 & 66.7 & 25.0
        & 83.3 & 0.0 & 0.0 & 0.0
        \\
        & Hard
        & 0.0 & 25.0 & 25.0 & 33.3
        & 75.0 & 0.0 & 0.0 & 0.0
        \\

        \addlinespace[0.15ex]

        \multirow{3}{*}{\textbf{Goal2Skill}}
        & Easy
        & 88.5 & 73.1 & 69.2 & 96.2
        & 80.8 & 3.8 & 46.2 & 11.5
        \\
        & Medium
        & 75.0 & 58.3 & 41.7 & 91.7
        & 66.7 & 0.0 & 25.0 & 8.3
        \\
        & Hard
        & 50.0 & 33.3 & 25.0 & 75.0
        & 41.7 & 0.0 & 8.3 & 0.0
        \\

        \midrule

        \rowcolor{gray!15}
        \multicolumn{10}{@{}l}{\textit{Oracle Reference}} \\

        \multirow{3}{*}{GroundSG+Oracle}
        & Easy
        & 96.2 & 100.0 & 100.0 & 100.0
        & 92.3 & 11.5 & 100.0 & 76.9
        \\
        & Medium
        & 8.3 & 100.0 & 100.0 & 100.0
        & 91.7 & 8.3 & 100.0 & 25.0
        \\
        & Hard
        & 0.0 & 91.7 & 100.0 & 91.7
        & 83.3 & 8.3 & 100.0 & 75.0
        \\

        \bottomrule
    \end{tabular}%
    }
\end{table}

\paragraph{Difficulty-stratified RoboMME results.}
Tables~\ref{tab:robomme_difficulty_cp} and
\ref{tab:robomme_difficulty_ri} show how the methods behave as task
difficulty increases. Success rates generally decrease from Easy to Hard
episodes, but the change varies considerably across tasks.

Goal2Skill remains relatively stable on several tasks that require
persistent progress tracking. On \emph{PickXtimes}, its success rate
changes from 96.2\% on Easy episodes to 83.3\% on Hard episodes.
On \emph{VideoUnmask}, it decreases from 92.3\% to 66.7\%.
A similar pattern appears on \emph{VideoPlaceOrder}, where Goal2Skill
achieves 96.2\%, 91.7\%, and 75.0\% on Easy, Medium, and Hard episodes,
respectively. In comparison, the strongest non-oracle result on the Hard
split of \emph{VideoPlaceOrder} is 33.3\%.

The breakdown also reveals different strengths across methods.
MemER remains strong on \emph{PickHighlight} and \emph{MoveCube},
while FrameSamp performs well on several trajectory- and
imitation-oriented tasks. Goal2Skill is strongest when increasing
difficulty requires the system to preserve task state over a longer
sequence, rather than simply execute a more precise motion. The
difficulty results therefore reinforce the task-level pattern observed
in the aggregate evaluation.

\end{document}